\documentclass[journal]{IEEEtran}
\usepackage{indentfirst}
\usepackage{amsmath}
\usepackage{amssymb}
\usepackage{graphicx}
\usepackage[tight,footnotesize]{subfigure}
\usepackage{multirow}
\usepackage{setspace}
\usepackage{ulem}
\usepackage[numbers,sort&compress]{natbib}
\begin{document}
\title{Image Restoration using Total Variation with Overlapping Group Sparsity}
\author{Jun Liu, Ting-Zhu Huang, Ivan W. Selesnick, Xiao-Guang Lv, Po-Yu Chen
\thanks{ The  work  of  Jun Liu and Ting-Zhu Huang is supported by NSFC (61170311), 973 Program (2013CB329404), and Sichuan Province Sci. \& Tech. Research Project (2012GZX0080). The work of Xiao-Guang Lv is supported by Nature science foundation of Jiangsu Province (BK20131209).
J. Liu and T.-Z. Huang are with School of Mathematical Sciences, University of Electronic Science and Technology of China, Chengdu, Sichuan, China (email: junliucd@163.com, tingzhuhuang@126.com).

Ivan W. Selesnick and Po-Yu Chen are with Polytechnic Institute of New York University, Brooklyn, NY 11201, USA, (email: selesi@poly.edu, poyupaulchen@gmail.com)

X.-G. Lv is with School of Science, Huaihai Institute of Technology, Lianyungang, Jiangsu, China, (email: xiaoguanglv@126.com)


}} \maketitle



\begin{abstract}
Image restoration is one of the most fundamental issues in imaging science. Total variation (TV) regularization is widely used in image restoration problems for its capability to preserve edges. In the literature, however, it is also well known for producing staircase-like artifacts. Usually, the high-order total variation (HTV) regularizer is an good option except its over-smoothing property. In this work, we study a minimization problem where the objective includes an usual $l_2$ data-fidelity term and an overlapping group sparsity total variation regularizer which can avoid staircase effect and allow edges preserving in the restored image. We also proposed a fast algorithm for solving the corresponding minimization problem and compare our method with the state-of-the-art TV based methods and HTV based method. The numerical experiments illustrate the efficiency and effectiveness of the proposed method in terms of PSNR, relative error and computing time.
\end{abstract}


\IEEEpeerreviewmaketitle

\section{Introduction}
\IEEEPARstart{I}{mage} restoration is one of the most fundamental issues in imaging science and plays an important role in many mid-level and high-level image processing applications. On account of the imperfection of an imaging system, a recorded image may be inevitably degraded during the process of image capture, transmission, and storage. The image formation process is commonly modeled as the following linear system
\begin{equation}\label{eq1}
g = Hf + \eta,\quad H\in \mathbb{R}^{n^2\times n^2}, f,g\in \mathbb{R}^{n^2},
\end{equation}
where the vectors $f$ and $g$ represent the $n\times n$ true scene and observation whose column vectors are the successive $n^2$-vectors of $f$ and $g$, respectively. $\eta$ is Gaussian white noise with zero mean, and $H$ is a blurring matrix constructed from the discrete point spread function, together with the given boundary conditions.

It is well known that image restoration belongs to a general class of problems which are rigorously classified as ill-posed problems \cite{KV1990,TA1977}.  To tackle the ill-posed nature of the problem, regularization techniques are usually considered to obtain a stable and accurate solution. In other words, we seek to approximately recover $f$ by minimizing the following variational problem:
\begin{equation} \label{eq2}
\min_f \left\{\frac{1}{2}\|g-Hf\|_2^2+ \alpha \varphi(f)\right\},
\end{equation}
where $\left\Vert \cdot \right\Vert_2$ denotes the Euclidean norm, $\varphi$ is conventionally called a regularization functional, and $\alpha>0$ is referred to as a regularization parameter which controls the balance between fidelity and regularization terms in (\ref{eq2}).

How to choose a good functional $\varphi$ is an active area of research in imaging science. In the early 1960s,  D. L. Phillips \cite{Phi1962} and A. N. Tikhonov \cite{Tik1963} proposed the definition of $\varphi$ as an $l_2$-type norm (often called Tikhonov regularization in the literature), that is, $\varphi = \left\Vert Lf\right\Vert_2^2$ with $L$ an identity operator or difference operator. The functional $\varphi$ of this type has the advantage of simple calculations, however, it produces a smoothing effect on the restored image, i.e., it overly smoothes edges which are important features in human perception. Therefore, it is not a good choice since natural images have many edges. To overcome this shortcoming, Rudin, Osher and Fatemi \cite{ROF1992} proposed to replace the $l_2$-type norm with the total variation (TV) seminorm, that is, they set $\varphi(f)=\left\Vert \nabla f \right\Vert_1$. Then the corresponding minimization problem is
\begin{equation} \label{eq3}
\min_f \left\{\frac{1}{2}\|g-Hf\|_2^2+ \alpha \left\Vert \nabla f \right\Vert_1\right\},
\end{equation}
where $\left\Vert \nabla f\right\Vert_1=\sum\limits_{i,j=1}^n\left\Vert (\nabla f)_{i,j}\right\Vert$ and the discrete gradient operator $\nabla: \mathbb{R}^{n^2}\rightarrow \mathbb{R}^{2\times n^2}$ is defined by $(\nabla f)_{i,j}=((\nabla_x f)_{i,j},(\nabla_y f)_{i,j} )$ with
\[
(\nabla_x f)_{i,j} =\left\{\begin{array}{lll}  f_{i+1,j}-f_{i,j}&{\rm if}&i<n,\\
                                            f_{1,j}-f_{n,j}&{\rm if}&i=n,\end{array}\right.
\]
and
\[
(\nabla_y f)_{i,j} =\left\{\begin{array}{lll}  f_{i,j+1}-f_{i,j}&{\rm if}&j<n,\\
                                            f_{i,1}-f_{i,n}&{\rm if}&j=n,\end{array}\right.
\]
for $i,j=1,2,\cdots,n$ and $f_{i,j}$ refers to the $((j-1)n+i)$th entry of the vector $f$ (it is the $(i,j)$th pixel location of the $n\times n$ image, and this notation is valid throughout the paper unless otherwise specified).

The problem (\ref{eq3}) is commonly referred to as the ROF model. The TV is isotropic if the norm $\left\Vert \cdot \right\Vert$ is the Euclidean norm and anisotropic if 1-norm is defined. In this work, we only consider the isotropic case since the isotropic TV usually behaves better than the anisotropic version.

In the literature, many algorithms have been proposed for solving (\ref{eq3}). In case $H$ is the identity matrix, then the problem (\ref{eq3}) is referred to as the denoising problem. In the pioneering work \cite{ROF1992}, the authors proposed to employ a time marching scheme to solve the associated Euler-Lagrange equation of (\ref{eq3}). However, their method is very slow due to CFL stability constraints \cite{SM2000}.  Later, Vogel and Oman \cite{Vogel1996} proposed a lagged diffusivity fixed point method to solve the same Euler-Lagrange equation of (\ref{eq3}). In \cite{Chan1999}, Chan and Mulet proved this method had a global convergent property and was asymptotically faster than the explicit time marching scheme. Chambolle \cite{Cham2004} studied a dual formulation of the TV denoising problem and proposed a semi-implicit gradient descent algorithm to solve the resulting constrained optimization problem. He also proved his algorithm is globally convergent with a suitable step size. In \cite{GO2009}, Goldstein and Osher proposed the novel split Bregman iterative algorithm to deal with the artificial constraints, their method has several advantages such as fast convergence rate and stability, etc.

In \cite{CGM1999}, Chan, Golub and Mulet considered to apply Newton's method to solve the nonlinear primal-dual system of the system (\ref{eq3}) for image deblurring problem. Recently, Wang \textit{et al.} \cite{FTVd2008} proposed a fast total variation deconvolution (FTVd) method which used splitting technique and constructs an iterative procedure of alternately solving a pair of easy subproblems associated with an increasing sequence of penalty parameter values. Almost at the same time, Huang, Ng and Wen \cite{HNW2008} proposed a fast total variation (Fast-TV) minimization method by introducing an auxiliary variable to replace the true image $f$. Their methods belong to penalty methods from the perspective of optimization. In \cite{BT2009}, Beck and Teboulle studied a fast iterative shrinkage-thresholding algorithm (FISTA) which is a non-smooth variant of Nesterov's optimal  gradient-based algorithm for smooth convex problems \cite{Nes2004}. Later, Afonso \textit{et al.} \cite{ABF2010} proposed an augmented Lagrangian shrinkage algorithm (SALSA) which is an instance of the so-called alternating direction method of multipliers (ADMM). More recently, Chan, Tao and Yuan \cite{CTY2013} proposed an efficient and effective method by imposing box constraint on the ROF model (\ref{eq3}). Their numerical experiments showed that their method could obtain much more accurate solutions and was superior to other state-of-the-art methods. The methods of solving ROF model (\ref{eq3}) mentioned above are just a few examples, we refer the interested readers to \cite{ZC2008,WT2010} and the references therein for further details.

Although total variation regularization has been proven to be extremely useful in a variety of applications, it is well known that TV yields staircase artifacts \cite{DS1996,CMM2000}. Therefore, the approaches involving the classical TV regularization often develop false edges that do not exist in the true image since they tend to transform smooth regions (ramps) into piecewise constant regions (stairs). To avoid these drawbacks, nonlocal methods were considered in \cite{BCM2005,ZBBO2010}. Besides, in the literature, there is a growing interest for replacing the TV regularizer by the high-order total variation (HTV) regularizer, which can comprise more than merely piecewise constant regions. The majority of the high-order norms involve second-order differential operators because piecewise-vanishing second-order derivatives lead to piecewise-linear solutions that better fit smooth intensity changes \cite{LBU2012}, namely, we choose the regularization functional $\varphi(f)=\left\Vert \nabla^2 f \right\Vert_1$. Then the minimization problem (\ref{eq2}) is treated as following HTV-based problem:
\begin{equation} \label{eq4}
\min_f \left\{\frac{1}{2}\|g-Hf\|_2^2+ \alpha \left\Vert \nabla^2 f \right\Vert_1\right\},
\end{equation}
where $\left\Vert \nabla^2 f \right\Vert_1 =\sum_{i=1}^n\left\Vert (\nabla^2 f)_{i,j}\right\Vert_2$ with $(\nabla^2 f)_{i,j} = \left((\nabla^2_{xx} f )_{i,j},(\nabla^2_{yx} f )_{i,j}; (\nabla^2_{xy} f )_{i,j}, (\nabla^2_{yy} f )_{i,j}\right)$. Note that $(\nabla^2_{st} f )_{i,j}$, $s,t\in \{x,y\}$ denotes the second order difference of $f$ at pixel $(i,j)$. The minimization problem (\ref{eq4}) is usually called LLT model which was first proposed by Lysaker, Lundervold, and Tai \cite{LLT2003}.

In \cite{LLT2003}, the authors applied gradient descent algorithm to solve the corresponding fourth-order partial differential equation. Later in \cite{CST2009}, Chen, Song and Tai employed the dual algorithm of Chambolle for solving (\ref{eq4}) and they verified that their method was faster than the original gradient descent algorithm. A similar dual method was also proposed by Steidl \cite{Steidl2006} but from the linear algebra point of view by consequently using matrix-vector notation. Recently, Wu and Tai considered to employ the alternating direction method of multipliers (ADMM) to tackle the problem (\ref{eq3}). Also, some other high order models have been proposed in the literature, we refer the interested reader to see \cite{CMM2000,YK2000,LT2006,HJ2012,BKP2010,PS2012} and references therein for details.

Note that there exist other different types of regularization functionals, such as the Markov random field (MRF) regularization \cite{GG1984}, the Mumford-Shah regularization \cite{MS1989}, and frame-based $l_1$ regularization \cite{COS2009}. In this paper, however, we consider to set $\varphi$ in (\ref{eq2}) to be the overlapping group sparsity total variation (OGS-TV) functional which we have introduced in \cite{SC2013} for the one-dimension signal denoising problem. The numerical experiments there showed that the OGS-TV regularizer can alleviate staircase effect effectively. Then it is natural to extend this idea to the 2-dimensional case such as image restoration considered in this work.

The rest of the paper is organized as follows. In the next section, we will briefly introduce the definition of the overlapping group sparsity total variation functional for image restoration. We will also review the majorization-minimization (MM) methods and ADMM, which are the essential tools for us to propose our efficient method. In section 3, we derive an efficient algorithm for solving the considered minimization problem. Consequently, in section 4, we give a number of numerical experiments of image denoising and image deblurring to demonstrate the effectiveness of the proposed method, as compared to some other state-of-the-art methods. Finally, discussions and conclusions are made in section 5.

\section{Preliminaries}
\subsection{OGS-TV}
In \cite{SC2013}, we have denoted a $K$-point group of the vector $s \in \mathbb{R}^n$ by
\begin{equation}\label{eq5}
s_{i,K} = [s(i),...,s(i+K-1)] \in \mathbb{R}^K
\end{equation}
Note that $s_{i,K}$ can be seen as a block of $K$ contiguous samples of $s$ staring at index $i$. With the notation (\ref{eq5}), a group sparsity regularizer \cite{FB2011,PF2011} is defined as
\begin{equation}\label{eq6}
\xi(s) = \sum_{i=1}^n \left\Vert s_{i,K}\right\Vert_2.
\end{equation}
The group size is denoted by $K$. For the two-dimensional case, we define a $K\times K$-point group of the image $f\in \mathbb{R}^{n^2}$ (note that the vector $f$ is obtained by stacking the $n$ columns of the $n\times n$ matrix)
\begin{equation}\label{eq7}
 \begin{split}
&\tilde{f}_{i,j,K} = \\
& \left[\begin{array}{cccc}  f_{i-m_1,j-m_1}  &f_{i-m_1,j-m_1+1}  &\cdots &f_{i-m_1,j+m_2}\\
                                         f_{i-m_1+1,j-m_1}  &f_{i-m_1+1,j-m_1+1}  &\cdots &f_{i-m_1+1,j+m_2}\\
                                          \vdots    &\vdots       &\ddots &\vdots\\
                                         f_{i+m_2,j-m_1}&f_{i+m_2,j-m_1+1}&\cdots &f_{i+m_2,j+m_2}\\
                                         \end{array}\right]\\
&\in \mathbb{R}^{K\times K}\\
 \end{split}
\end{equation}
with $m_1 = [\frac{K-1}{2}], m_2 = [\frac{K}{2}]$, where $[x]$ denotes the greatest integer not greater than $x$. Let ${f}_{i,j,K}$ be a vector which is obtained by stacking the $K$ columns of the matrix $\tilde{f}_{i,j,K}$, i.e., ${f}_{i,j,K}=\tilde{f}_{i,j,K}(:)$. Then the overlapping group sparsity functional of a two-dimensional array can be defined by
\begin{equation}\label{eq8}
 \phi(f) = \sum_{i,j=1}^n \left\Vert f_{i,j,K}\right\Vert_2.
\end{equation}
The group size of functional (\ref{eq8}) is denoted by $K\times K$. Consequently, we set the regularization functional $\varphi$ in (\ref{eq2}) to be of the form
\begin{equation}\label{eq9}
 \varphi(f) = \phi(\nabla_x f) + \phi(\nabla_y f).
\end{equation}

In (\ref{eq9}), if $K=1$, then $\varphi(f)$ is the commonly used anisotropic TV functional. Then we refer to the regularizer $\varphi$ in (\ref{eq9}) as the overlapping group sparsity anisotropic total variation functional (OGS-ATV).

\subsection{ADMM}
The ADMM technique was initially proposed to solve the following constrained separable convex optimization problem:
\begin{equation}\label{cop}
\begin{array}{ll}
  \min  & \theta_1(x_1)+\theta_2(x_2) \\
 {\rm s.\ t} & A_1x_1+  A_2x_2=d,\\
   & x_i\in\mathcal {X}_i, i=1,2
\end{array}
\end{equation}
where $\theta_i: \mathcal {X}_i\rightarrow\mathbb{R}$ are closed convex functions, $A_i\in\mathbb{R}^{l\times m_i}$ are linear transforms, $\mathcal{X}_i\in \mathbb{R}^{m_i}$ are nonempty closed convex sets, and $d\in\mathbb{R}^{l}$ is a given vector.

Using a Lagrangian multiplier $\lambda \in\mathbb{R}^{l}$ to the linear constraint in (\ref{cop}), the augmented Lagrangian function \cite{Hes1969} for problem (\ref{cop}) is
\begin{equation}\label{alf}
\begin{array}{cl}
\mathcal{L}(x_1,x_2,\lambda) &=  \theta_1(x_1)+\theta_2(x_2)+\lambda^T(A_1x_1+  A_2x_2-d)\\
   & +\frac{\sigma}{2}\left\Vert A_1x_1+  A_2x_2-d\right\Vert_2^2
\end{array}
\end{equation}
where $\lambda\in\mathbb{R}^{l}$ is the Lagrange multiplier and $\sigma$ is a penalty parameter, which controls the linear constraint. The idea of the ADMM is to find a saddle point $(x_1^*,x_2^*,\lambda^*)$ of $\mathcal{L}$. Usually, the ADMM consists in minimizing $\mathcal{L}(x_1,x_2,\lambda)$ alternatively, subject to $x_1,x_2,\lambda$, such as minimizing $\mathcal{L}$ with respect to $x_1$, keeping $x_2$ and $\lambda$ fixed. Notice that the term $\lambda^T(A_1x_1+  A_2x_2-d)+\frac{\sigma}{2}\left\Vert A_1x_1+  A_2x_2-d\right\Vert_2^2$ in the definition of the augmented Lagrangian functional $\mathcal{L}(x_1,x_2,\lambda)$ in (\ref{alf}) can be written as a single quadratic term after simple mathematical operations, leading to the following alternative form for a simple but powerful algorithm: the ADMM
\\\\
{\it
\begin{tabular}{l}
\hline
\hline
{\textup{\textbf{Algorithm 1}}} \textup{ADMM for the minimization problem (\ref{cop})}
\\ \hline
\textit{\textbf{initialization}}: Starting point $(x_1^0,x_2^0,\lambda^0)$, $\sigma>0$,
\\
\textit{\textbf{iteration}}:\\
$x_1^{k+1}=\arg\min_{x_1} \theta_1(x_1)+\frac{\sigma}{2}\left\Vert A_1x_1+ A_2x_2^k-d + b^k\right\Vert_2^2;$ \\
$x_2^{k+1}=\arg\min_{x_2}\theta_2(x_1)+\frac{\sigma}{2}\left\Vert A_1x_1^k+ A_2x_2-d + b^k\right\Vert_2^2;$ \\
$b^{k+1}=b^k+\sigma(A_1x_1^{k+1}+ A_2x_2^{k+1}-d);$\\
$k = k+1;$\\
\textbf{until a stopping criterion is satisfied.}\\
\hline
\end{tabular}}
\\
\\

An important advantage of the ADMM is to make full use of the separable structure of the objective function $\theta_1(x_1)+\theta_2(x_2)$. Note that the ADMM is a splitting version of the augmented Lagrangian method where the augmented Lagrangian method's subproblem is decomposed into two subproblems in the Gauss-Seidel fashion at each iteration, and thus the variables $x_1$ and $x_2$ can be solved separably in alternating order. The convergence of the alternating direction method can be found in \cite{EB1992,HY1998}. Moreover, we have
\begin{equation}\label{conv1}
\begin{split}
\left\|\begin{array}{c}
    A_2(x_2^{k+1}-x_2^*) \\
    (b^{k+1}-b^*)
  \end{array}
\right\|^2_2&\leq
 \left\|\begin{array}{c}
     A_2(x_2^{k}-x_2^*) \\
     (b^{k}-b^*)
  \end{array}
\right\|^2_2\\
&\quad -\left\|\begin{array}{c}
                    A_2(x_2^{k+1}-x_2^k) \\
                    (b^{k+1}-b^k)
                    \end{array}
             \right\|^2_2.
\end{split}
\end{equation}
Hence, $\lim\limits_{k\rightarrow \infty}b^k=b^*$ and $\lim\limits_{k\rightarrow \infty}A_2x_2^k=A_2x_2^*$. Especially, if matrices $A_1$ and $A_2$ have full column rank, it leads to $\lim\limits_{k\rightarrow \infty}x_1^k=x_1^*$ and $\lim\limits_{k\rightarrow \infty}x_2^k=x_2^*$.

\subsection{MM}

The MM method substitutes a simple optimization problem for a difficult optimization problem. That is to say, instead of minimizing a difficult cost functional $R(f)$ directly, the MM approach solves a sequence of optimization problems, $Q(f,f^k), k = 0, 1, 2, \cdots$. The idea is that each $Q(f,f^k)$ is easier to solve that $R(f)$. Of course, iteration is the price we pay for simplifying the original problem. Generally, a MM iterative algorithm for minimizing $R(f)$ has the form
\begin{equation}\label{eq13}
  f^{k+1} = \arg\min_f Q(f,f^k)
\end{equation}
where $Q(f,f')\geq R(f)$, for any $f,f'$, and $Q(f^k,f^k)= R(f^k)$, i.e., each functional $Q(f,f')$  is a majorizor of $R(f)$. When $R(f)$ is convex, then under mild conditions, the sequence $f^k$ produced by (\ref{eq13}) converges to the minimizer of $R(f)$.

A good majorizing functional $Q$ usually satisfies the following characteristics \cite{HL2004}: (a) avoiding large matrix inversions, (b) linearizing an optimization problem, (c) separating the parameters of an optimization problem, (d) dealing with equality and inequality constraints gracefully, or (e) turning a nondifferentiable problem into a smooth problem. More details about the MM procedure can be found in  \cite{HL2004,FBN2007} and the references therein.

Before we proceed with the discussion of the proposed method, we consider a minimization problem of the form
\begin{equation}\label{eq14} 
\min_v \left\{R(v)=\frac{1}{2}\left\|v-v_0\right\|_2^2 + \mu \phi(v)\right\}, v\in \mathbb{R}^{n^2}
\end{equation}
where $\mu$ is a positive parameter and the functional $\phi$ is given by (\ref{eq8}). In \cite{CS2014}, we analysed this problem elaborately. However, for the sake of completeness, we briefly introduce the solving method here. To derive an effective and efficient algorithm with the MM approach for solving the problem (\ref{eq14}), we need a majorizor of $R(v)$, and fortunately, we only need to find a majorizor of $\phi(v)$ because of the simple quadratic term of the first term in (\ref{eq14}). To this end, note that
\begin{equation}\label{eq15} 
\frac{1}{2\|u\|_2}\|v\|_2^2 + \frac{1}{2}\|u\|_2\geq\|v\|_2
\end{equation}
for all $v$ and $u\neq 0$ with equality when $u=v$. Substituting each group of $\phi(v)$ into (\ref{eq15}) and summing them, we get a majorizor of $\phi(v)$
\begin{equation}\label{eq16} 
\begin{split}
&P(v,u)\\
&=\frac{1}{2}\sum_{i,j=1}^n \left[\frac{1}{\left\|u_{i,j,K}\right\|_2}\left\|v_{i,j,K}\right\|_2^2+\left\|u_{i,j,K}\right\|_2\right]
\end{split}
\end{equation}
with
\begin{equation}\label{eq17}
P(v,u)\geq \phi(v),\quad P(u,u) = \phi(u)
\end{equation}
provided $\left\|u_{i,j,K}\right\|_2\neq 0$ for all $i,j$. With a simple calculation, $P(v,u)$ can be rewritten as
\begin{equation}\label{eq18}
P(v,u) = \frac{1}{2}\left\|\Lambda(u)v\right\|_2^2 + C,
\end{equation}
where $C$ is a constant that does not depend on $v$, and $\Lambda(u)$ is a diagonal matrix with each diagonal component
\begin{equation}\label{eq19}
\left[\Lambda(u)\right]_{l,l} = \sqrt{\sum_{i,j=-m_1}^{m_2} \left[\sum_{k_1,k_2=-m_1}^{m_2}\left|u_{r-i+k_1,t-j+k_2}\right|^2 \right]^{-\frac{1}{2}}}
\end{equation}
with $l=(t-1)n+r, r,t = 1,2,\cdots,n$. The entries of $\Lambda$ can be easily computed using Matlab built-in function \verb"conv2". Then a majorizor of $R(v)$ can be easily given by
\begin{equation}\label{eq20}
\begin{split}
Q(v,u) &= \frac{1}{2}\left\|v-v_0\right\|_2^2 + \mu P(v,u)\\
       &=\frac{1}{2}\left\|v-v_0\right\|_2^2+
\frac{\mu}{2}\left\|\Lambda(u)v\right\|_2^2 + \mu C
\end{split}
\end{equation}
with $Q(v,u)\geq R(v)$ for all $u, v$, and $Q(u,u)=R(u)$. To minimize $R(v)$, the MM aims to iteratively solve
\begin{equation}\label{eq21}
v^{k+1} = \arg\min_v \frac{1}{2}\left\|v-v_0\right\|_2^2+
\frac{\mu}{2}\left\|\Lambda(v^k)v\right\|_2^2, k = 0,1,\cdots.
\end{equation}
which has the solution
\begin{equation}\label{eq22}
v^{k+1} = \left(I+\mu \Lambda(v^k)^T\Lambda(v^k)\right)^{-1}v_0,
\end{equation}
where $I$ is a identity matrix with the same size of $\Lambda(v^k)$. Note that the inversion of the matrix $\left(I+\mu \Lambda(v^k)^T\Lambda(v^k)\right)$ can be computed very efficiently via simple component-wise calculation. To summerize, we obtain the Algorithm 2 for solving the problem (\ref{eq14}).
\\\\
{\it
\begin{tabular}{l}
\hline
\hline
{\textup{\textbf{Algorithm 2}}} \\ \textup{for solving the minimization problem (\ref{eq14})}
\\ \hline
\textit{\textbf{initialization}}:\\
{\rm Starting point} $v^0=v_0$, $k=0,\mu$, $K$, {\rm Maximum inner}\\
{\rm iterations} $Nit$.\\
\textit{\textbf{iteration}}:\\
\textup{1}.\ $\left[\Lambda(u)\right]_{l,l}= \sqrt{\sum\limits_{i,j=-m_1}^{m_2} \left[\sum\limits_{k_1,k_2=-m_1}^{m_2}\left|u_{r-i+k_1,t-j+k_2}\right|^2 \right]^{-\frac{1}{2}}}$\\
\textup{2}.\ $v^{k+1} = \left(I+\mu \Lambda(v^k)^T\Lambda(v^k)\right)^{-1}v_0$\\
\textup{3}.\ $k=k+1$;\\
\textbf{until} $\|v^{k+1}-v^k\|_2/\|v^{k}\|_2<\epsilon_l$ or $k<N$.\\
\hline
\end{tabular}}
\\
\\
\section{Proposed algorithm}
With the defiNion of (\ref{eq9}), in this section, we address the minimization problem of the form
\begin{equation}\label{eq23} 
\min_f \left\{\frac{1}{2}\|g-Hf\|_2^2+ \alpha \left(\phi((\nabla_x f)) + \phi((\nabla_y f))\right)\right\}.
\end{equation}
We refer to this model as $L_2$-OGS-ATV.
Note that for any true digital image, its pixel can attain only a fiNe number of values. Hence, it is natural to require all pixel values of the restored image to lie in a certain interval $[a_l, a_u]$. Such a constraint is called the box constraint \cite{CM2012}. For instance, the images considered in this work are all 8-bit images, we would like to restore them in a dynamic range $[0, 255]$. For convenience, we define an orthogonal projection operator $\mathcal{P}_\Omega$ on the set $\Omega=[a_l, a_u]$,

\begin{equation}\label{eq24}
\mathcal{P}_\Omega(f)_{i,j} =
\left\{ \begin{aligned}
         &a_l,      & f_{i,j}<a_l\quad\quad \\
         &f_{i,j},&\ \ f_{i,j}\in [a_l\ a_u]\\
         &a_u,    & f_{i,j}>a_u\quad  \ \ \\
         \end{aligned}
\right.
\end{equation}

By introducing new auxiliary variables $v_x,v_y,z$, we change the minimization problem (\ref{eq23}) together with a constraint (\ref{eq24}) to the equivalent constrained minimization problem

\begin{equation}\label{eq25}
\begin{array}{ll}
\min\limits_f &\left\{\frac{1}{2}\|g-Hf\|_2^2+ \alpha \left(\phi(v_x) + \phi(v_y)\right)+\mathcal{P}_\Omega(z)\right\}\\
\ {\rm s.t} & v_x=\nabla_x f, v_y=\nabla_y f, z = f
\end{array}
\end{equation}

Thus, problem (\ref{eq25}) satisfies the framework in (\ref{cop}) with the following specifications:
\\
1) $x_1: = f, x_2: = \begin{pmatrix}
    v_x  \\
     v_y \\
     z \\
  \end{pmatrix}, \mathcal{X}_1 = \mathbb{R}^{n^2}, \mathcal{X}_2 = \mathbb{R}^{3n^2}$;
\\
2) $\theta_1(x_1):=\frac{1}{2}\|g-Hf\|_2^2, \theta_2(x_2):=\alpha\left(\phi((v_x)) + \phi((v_y))\right)+\mathcal{P}_\Omega(z)$;
\\
3) \[
A_1=
  \begin{pmatrix}
    \nabla_x  \\
     \nabla_y \\
     I \\
  \end{pmatrix},
A_2 =
  \begin{pmatrix}
    -I &    &  \\
       & -I &  \\
       &    & -I \\
  \end{pmatrix},
d=
  \begin{pmatrix}
    0 \\
    0 \\
    0 \\
  \end{pmatrix};
\]

According to Algorithm 1, we get the iterative scheme
\begin{equation}\label{eq26}
\begin{split}
f^{k+1} = &\arg\min_f \frac{1}{2}\|g-Hf\|_2^2
+ \frac{\sigma}{2}\left\{\left\|\nabla_x f-v_x^k +b_1^k\right\|_2^2\right.\\
&+\left.\left\|\nabla_y f-v_y^k +b_2^k\right\|_2^2+\left\|f -z^k  +b_3^k\right\|_2^2\right\}
\end{split}
\end{equation}

\begin{equation}\label{eq27}
\begin{aligned}
  \begin{pmatrix}
    v_x^{k+1}  \\
     v_y^{k+1} \\
     z^{k+1} \\
  \end{pmatrix}& = \arg\min_{v_x,v_y,z} \frac{\sigma}{2}\left\{\left\|\nabla_x f^{k+1}-v_x+b_1^k\right\|_2^2\right.\\
&+\left.\left\|\nabla_y f^{k+1}-v_y+b_2^k\right\|_2^2+\left\|f^{k+1} -z+b_3^k\right\|_2^2\right\}\\
&+\alpha \left(\phi(v_x) + \phi(v_y)\right)+\mathcal{P}_\Omega(z)
\end{aligned}
\end{equation}

\begin{equation}\label{eq28}
  \begin{pmatrix}
    b_1^{k+1}  \\
     b_2^{k+1} \\
     b_3^{k+1} \\
  \end{pmatrix}=
  \begin{pmatrix}
    b_1^{k} + \sigma\left(\nabla_x f^{k+1} -v_x^{k+1}\right)  \\
     b_2^{k}+\sigma\left(\nabla_y f^{k+1} -v_y^{k+1}\right) \\
     b_3^{k} + \sigma\left(f^{k+1}-z^{k+1}\right)\quad\  \\
  \end{pmatrix}\qquad \qquad\qquad\ \
\end{equation}

We now investigate these subproblems one by one.  The minimization problem (\ref{eq26}) with respect to $f$ is a least square problem which is equivalent to the corresponding normal equation

\begin{equation}\label{eq29}
\begin{array}{l}
  (H^TH+\sigma\nabla_x^T\nabla_x+\sigma \nabla_y^T \nabla_y+\sigma I)f^{k+1}= \\
\left(H^T g + \sigma\nabla_x^T (v_x^k-b_1^k)+\sigma\nabla_y^T (v_y^k-b_2^k)+\sigma(z^k-b_3^k)\right).
\end{array}
\end{equation}

Since the parameters $\sigma$ is positive, the coefficient matrix in (\ref{eq29}) is always invertible even when $H^TH$ is singular. Note that $H,\nabla_x$, and $\nabla_y$ block circulant with circulant blocks (BCCB) when the periodic boundary conditions are used.  We know that the computations with BCCB matrices can be very efficiently performed by using fast Fourier transforms (FFTs).

Clearly, the minimization (\ref{eq27}) with respect to $v_x,v_y,z$ are decoupled, i.e., they can be solved separately. Considering $v_x$, we have
\begin{equation}\label{eq30}
v_x^{k+1}=\arg\min_{v_x} \frac{\sigma}{2} \left\|v_x-\left(\nabla_x f^{k+1}+b_1^k\right)\right\|_2^2+  \alpha \phi(v_x)
\end{equation}
\\
The $v_y-$subproblem corresponds to the following optimization problem
\begin{equation}\label{eq31}
v_y^{k+1}=\arg\min_{v_y} \frac{\sigma}{2} \left\|v_y-\left(\nabla_y f^{k+1}+b_2^k\right)\right\|_2^2+  \alpha \phi(v_y)
\end{equation}
For simplicity, we denote $\beta = \alpha/\sigma$. It can be observed that problems (\ref{eq30}) and (\ref{eq31}) match the framework of the problem (\ref{eq14}), thus the solutions of (\ref{eq30}) and (\ref{eq31}) can be obtained by using Algorithm 3, respectively.
\begin{figure}[ht]
  \centering
    \includegraphics[width=0.45\textwidth]{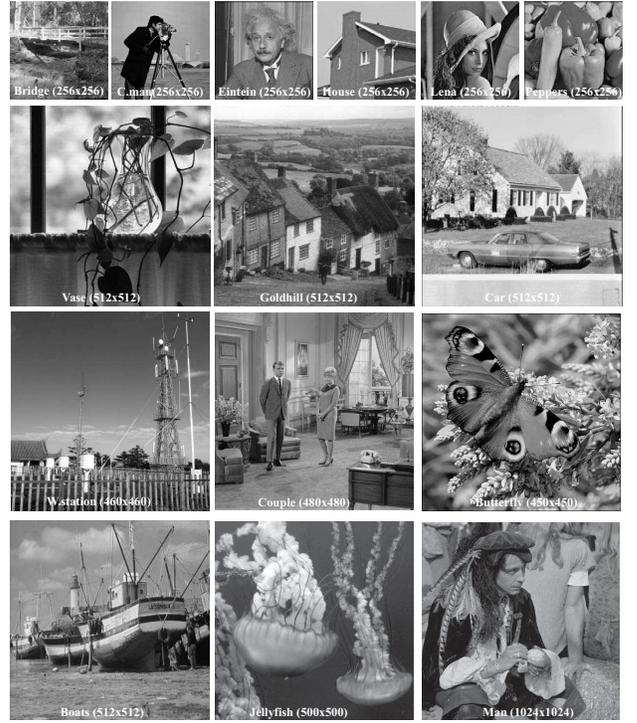}
    \caption{Different test images for the numerical experiments; the size of images varies from $256\times256$ to $1024\times1024$.}
\end{figure}
Besides, the $z-$subproblem can be solved by the simple projection $\mathcal{P}_\Omega$ onto the box
\begin{equation}\label{eq32}
z^{k+1} = \mathcal{P}_\Omega\left[f^{k+1}+b_3^k\right]
\end{equation}

Based on the discussions above, we get the resulting algorithm for solving (\ref{eq23}) shown as Algorithm 3.
\\\\
{\it
\begin{tabular}{l}
\hline
\hline
{\textup{\textbf{Algorithm 3}}} \\ \textup{OGSATV-ADM4 for the minimization problem (\ref{eq23})}
\\ \hline
\textit{\textbf{initialization}}: \\
{\rm Starting point} $v_x=v_y=g$, $k=0$, $\beta>0,\sigma>0,K$,\\$b_i^k$=0, $i=1,2,3$ {\rm Maximum inner iterations} $MaxIter$.\\
\textit{\textbf{iteration}}:\\
\textup{1}.\ Compute $f^{k+1}$ according to $(\ref{eq29})$\\
\textup{2}.\ Compute $v_x^{k+1}$ according to $(\ref{eq30})$ \\
\textup{3}.\ Compute $v_y^{k+1}$ according to $(\ref{eq31})$ \\
\textup{4}.\ Compute $z^{k+1}$ according to $(\ref{eq32})$ \\
\textup{5}.\ Update $b_i^{k+1},i=1,2,3$ according to $(\ref{eq28})$ \\
\textup{6}.\ k=k+1;\\
\textbf{until a stopping criterion is satisfied.}\\
\hline
\end{tabular}}
\\
\\
\\

Obviously, OGSATV-ADM4 is an instance of ADMM if the minimizations in steps 1$\sim$4 are solved exactly (i.e., the subproblems have closed-form solutions), the convergence of OGSATV-ADM4 is guaranteed. Note that, although steps (2) and (3) in Algorithm 3 can not be solved exactly, the convergence of Algorithm 3 is not compromised as long as the sequence of errors of successive solutions in (2) and (3) are absolutely summable, respectively. The corresponding theoretical proof is given elaborately in \cite{EB1992} and we will also verify this property in our numerical experiments.

\section{Numerical results}
In this section, we present some numerical results to illustrate the performance of the proposed method. The test images are shown in Fig. 1 with sizes from $256\times 256$ to $1024\times 1024$. All experiments are carried out on Windows 7 32-bit and Matlab v7.10 running on a desktop equipped with an Intel Core i3-2130 CPU 3.4 GHz and 4 GB of RAM.

\begin{figure}[htbp]
  \centering
  \subfigure[]{
    \label{fig2:subfig:a}
    \includegraphics[width=0.45\textwidth]{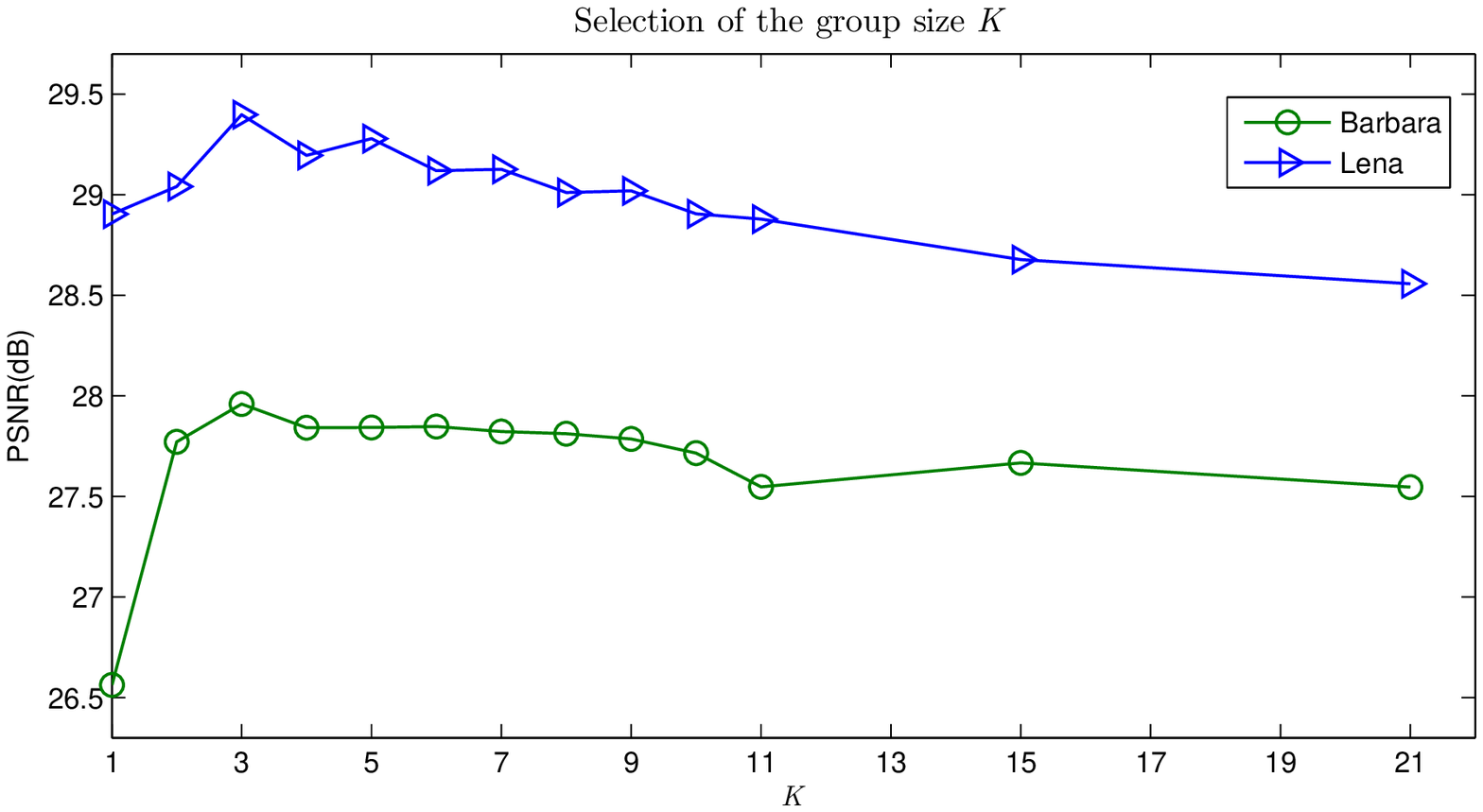}}\\
  \subfigure[]{
    \label{fig2:subfig:b}
    \includegraphics[width=0.45\textwidth]{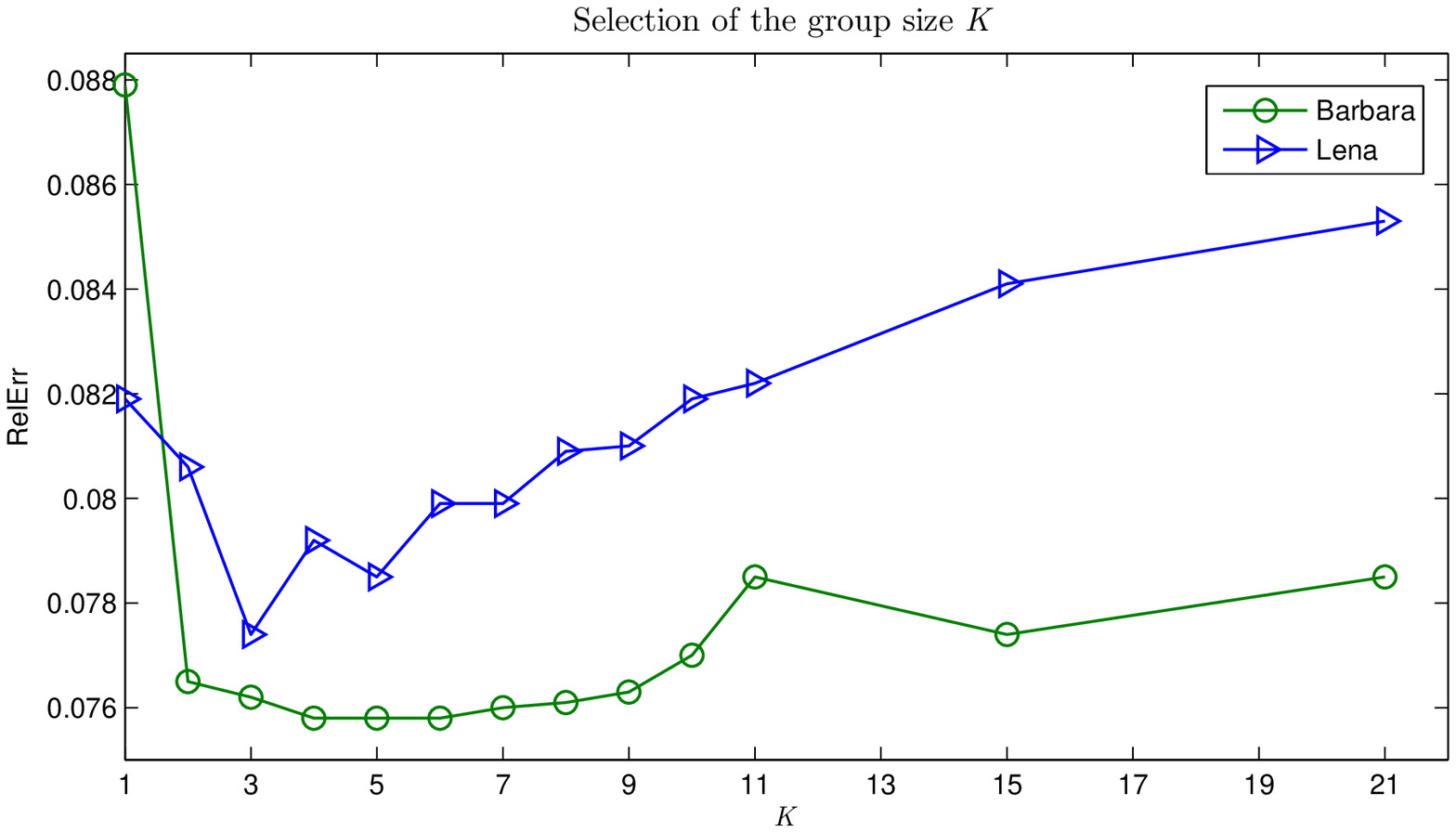}}
    \caption{PSNRs and RelErrs for images restored by OGSATV-ADM4 with different group sizes $K$
             }
\end{figure}

\begin{table}[htbp]
\begin{center}
\setlength{\abovecaptionskip}{0pt} \setlength{\belowcaptionskip}{12pt}
\centering
\caption{Restoration results for different numbers ($N$) of MM iterations in the OGSATV-ADM4}
\centering
    \begin{tabular}{|c|c|c|c|c|c|}
    \hline
          & $N$   & PSNR (dB) & RelErr & Iter  & Time (s) \\
    \hline
    \multirow{5}{*}{Lena} & 1     & 28.83 & 0.0827 & 16    & 0.5 \\
\cline{2-6}          & 5     & 29.40 & 0.0774 & 23    & 1.2 \\
\cline{2-6}          & 20    & 29.45 & 0.0769 & 26    & 3.1 \\
\cline{2-6}          & 200   & 29.45 & 0.0769 & 36    & 35.5 \\
\cline{2-6}          & 1000  & 29.45 & 0.0769 & 36    & 199.1 \\
    \hline
    \multirow{5}{*}{Barbara} & 1     & 27.49 & 0.0790 & 9     & 0.3 \\
\cline{2-6}          & 5     & 27.80  & 0.0762 & 19    & 1.0 \\
\cline{2-6}          & 20    & 27.82 & 0.0761 & 20    & 2.5 \\
\cline{2-6}          & 200   & 27.82 & 0.0761 & 20    & 20.4 \\
\cline{2-6}          & 1000  & 27.82 & 0.0761 & 20    & 119.2 \\
    \hline
    \end{tabular}%
  \label{tab1:addlabel}%
 \end{center}
\end{table}%

The quality of the restoration results is measured quantitatively by using the relative error (RelErr) and the peak signal-to-noise ratio (PSNR). The higher PSNR value, the higher image quality. We also use blur signal-to-noise ratio (BSNR) to describe how much noise is added in the blurry image. Suppose $f,g,\tilde{f}$, and $\eta$ are the original image, the observed image, the restored image, and the noise, respectively. The relative error of the restored image with respect to the original image is defined by:

\begin{equation}
\textmd{RelErr} = \frac{\|u - \hat{u}\|_2}{\|u\|_2}
\end{equation}
\\
The PSNR is defined as follows:
\begin{equation}
\textrm{PSNR} = 10\textmd{log}\frac{n^2\textmd{Max}_f^2}{\|f-\tilde{f}\|_2^2},
\end{equation}
where $\textmd{Max}_f$ is the maximum possible pixel value of the image $f$, such as, when the pixels are represented by using 8 bits per sample, it is 255. The BSNR is given by
\begin{equation}
\textrm{BSNR} = 20\log_{10}\frac{\|g\|_2}{\|\eta\|_2}
\end{equation}

The stopping criterion used in this work is set to be
\begin{equation}\label{eq36}
\frac{\| \mathcal{J}_{k+1} - \mathcal{J}_{k}\|}{\|\mathcal{J}_{k}\|} \leq \epsilon,
\end{equation}
where $\mathcal{J}_k$ is the objective function value of the corresponding model in the $k$-th iteration, we set $\epsilon = 1\times 10^{-5}$ for all our tests.
\begin{table*}
\setlength{\abovecaptionskip}{0pt} \setlength{\belowcaptionskip}{12pt}
\centering
\caption{COMPARISON OF THE PERFORMANCE OF FOUR METHODS WITH DIFFERENT NOISE LEVEL}
    \begin{tabular}{|c|r|c|c|c|c|}
    \hline
          &       & SplitBregman & Chambolle & LLT-ALM & OGSATV-ADM4 \\
    \hline
    $\sigma$ & \multicolumn{1}{c|}{Image} & PSNR/RelErr/Time/Iter & PSNR/RelErr/Time/Iter & PSNR/RelErr/Time/Iter & PSNR/RelErr/Time/Iter \\
    \hline
    \multirow{9}{*}{15} & \multicolumn{1}{c|}{Einstein} & 31.00/0.063/0.30/22  & 31.02/0.063/0.80/80  & 30.89/0.064/0.72/27 & 31.37/0.061/0.94/23  \\
\cline{2-6}          & \multicolumn{1}{c|}{Lena} & 30.48/0.068/0.26/20  & 30.52/0.068/0.72/75 & 30.40/0.069/0.59/22 & 30.83/0.066/0.68/19 \\
\cline{2-6}          & \multicolumn{1}{c|}{W.station} & 30.61/0.061/1.10/42 & 30.60/0.061/3.54/77 & 30.39/0.062/2.09/18  & 31.12/0.057/3.64/26 \\
\cline{2-6}          & \multicolumn{1}{c|}{Couple} & 31.09/0.056/0.64/19  & 31.10/0.056/4.36/82 & 31.33/0.055/5.95/44 & 31.57/0.053/2.67/19 \\
\cline{2-6}          & \multicolumn{1}{c|}{Boats} & 30.53/0.055/2.75/39  & 30.54/0.055/4.68/78 & 30.69/0.054/2.25/17  & 30.99/0.052/3.05/19 \\
\cline{2-6}          & \multicolumn{1}{c|}{Car} & 30.49/0.046/1.20/18  & 30.50/0.046/4.45/75 & 30.80/0.044/3.76/28  & 31.02/0.043/3.80/23 \\
\cline{2-6}          & \multicolumn{1}{c|}{Vase} & 31.32/0.051/1.44/21  & 31.29/0.051/4.75/80  & 31.52/0.049/3.05/23 & 31.75/0.048/4.23/26  \\
\cline{2-6}          & \multicolumn{1}{c|}{Goldhill} & 30.66/0.059/2.72/39  & 30.66/0.059/4.56/77  & 30.90/0.057/2.76/21 & 31.14/0.055/3.04/19  \\
\cline{2-6}          & \multicolumn{1}{c|}{Man} & 31.15/0.067/5.39/20 & 31.15/0.067/19.66/81  & 31.39/0.065/10.25/20  & 31.54/0.064/12.65/19  \\
    \hline
    \hline
    \multirow{9}{*}{30} & \multicolumn{1}{c|}{Einstein} & 28.02/0.089/0.64/48  & 28.01/0.089/1.13/ 120  & 27.61/0.093/0.57/22 & 28.35/0.086/1.20/33  \\
\cline{2-6}          & \multicolumn{1}{c|}{Lena} & 27.21/0.010/0.59/43 & 27.22/0.100/1.03/112 & 27.00/0.102/0.55/21  & 27.47/0.097/1.24/34  \\
\cline{2-6}          & \multicolumn{1}{c|}{W.station} & 26.99/0.092/1.15/43  & 27.00/0.092/5.11/113  & 26.77/0.094/2.32/20  & 27.50/0.087/4.81/35  \\
\cline{2-6}          & \multicolumn{1}{c|}{Couple} & 28.12/0.079/1.46/48  & 28.12/0.079/6.30/121  & 28.17/0.079/3.16/27  & 28.50/0.076/5.03/34  \\
\cline{2-6}          & \multicolumn{1}{c|}{Boats1} & 27.46/0.078/3.29/46  &  27.47/0.078/7.43/117 & 27.51/0.078/3.05/22  & 27.86/0.075/6.24/34  \\
\cline{2-6}          & \multicolumn{1}{c|}{Car} & 27.21/0.067/3.29/45  & 27.22/0.067/7.01/113  &  27.17/0.067/3.38/24  & 27.68/0.063/5.64/34  \\
\cline{2-6}          & \multicolumn{1}{c|}{Vase} & 28.04/0.074/3.22/45 & 28.01/0.074/7.08/117  & 28.18/0.072/3.59/26 & 28.43/0.070/5.51/33 \\
\cline{2-6}          & \multicolumn{1}{c|}{Goldhill} & 27.76/0.082/3.21/45 & 27.77/0.082/7.14/117 & 27.83/0.081/3.57/26 & 28.18/0.078/5.74/33  \\
\cline{2-6}          & \multicolumn{1}{c|}{Man} & 28.25/0.093/13.36/47  & 28.24/0.093/29.53/120  & 28.41/0.091/13.75/26 & 28.60/0.089/21.70/32  \\
    \hline
    \end{tabular}%
  \label{tab2:addlabel}%
\end{table*}%

Before starting the comparisons of our method with other state-of-the-art methods, we first study the setting of the group window size $K$ and the inner iterations $N$ of Algorithm 3. Note that, OGSATV-ADM4 satisfies the framework of ADMM and hence converges for any penalty parameter $\sigma>0$; however, the choice of $\sigma>0$ does influence the speed of the algorithms. Unfortunately, there is no work on methods to choose this parameter for optimal speed \cite{FB2010}. We empirically found a satisfying rule of thumb for our method adopted in all experiments with $\sigma=\lambda/3$.

\subsection{Study on some parameters}
Two images ``Barbara" and ``Lena" are used to study the effect of different choices of the group window size $K$ and the inner iterations $N$ in this subsection. We first check how the group size $K$ impacts the performance of the proposed method. We choose $N=5$ for the experiments. The two images were blurred by Gaussian kernel with radius $3$ and standard deviation $\delta = 2$, and then contaminated by Gaussian noise with BSNR = 40. We plot the PSNR values for the restored images with best tuned regularization parameter $\lambda$. Then selection of $K$ varies from $1$ to $21$. Apparently, if $K=1$, the model (\ref{eq23}) is the classic anisotropic TV (ATV-L2) case. From Fig. 2, we observe that small $K$ such as $3$ gives satisfying results, then we empirically choose $K=3$ in the following experiments.

Next we consider restricting the number $N$ of MM method in a subproblem using MMGS-Skrg algorithm. In Table 1, we set the number $N$ of the MM iterations to be $1, 5, 20, 200$ or $1000$ for the restoration of the two images in the tests. Obviously, small (large) inner iteration $N$ leads to small (large) computational cost. In the table, except $N = 1$, we observe that the quality in terms of PSNR and RelErr of the restored images by proposed OGSATV-ADM4 is almost the same, but with significantly different computational times. These results show that even though steps (2) and (3) of OGSATV-ADM4 are not solved exactly, a quite small number of the MM iterations are sufficient to achieve satisfying results. Based on these observations, we set $N=5$ for our method.

\subsection{Comparison with other state-of-the-art methods}
In this section, we report the experimental results aimed at comparing the proposed OGSATV-ADM4 with the current state-of-the-art methods proposed in \cite{GO2009}, \cite{CTY2013}, \cite{WT2010} and \cite{HNW2008} for image restoration including image denoising and image deblurring. For the sake of a fair comparison, the proposed and reference methods have been terminated using the same stopping criterion (\ref{eq36}) with $\epsilon=1\times 10^{-5}$.
\begin{figure}[htbp]
  \centering
  \subfigure[Noisy image]{
    \label{fig3:subfig:a}
    \includegraphics[width=0.23\textwidth]{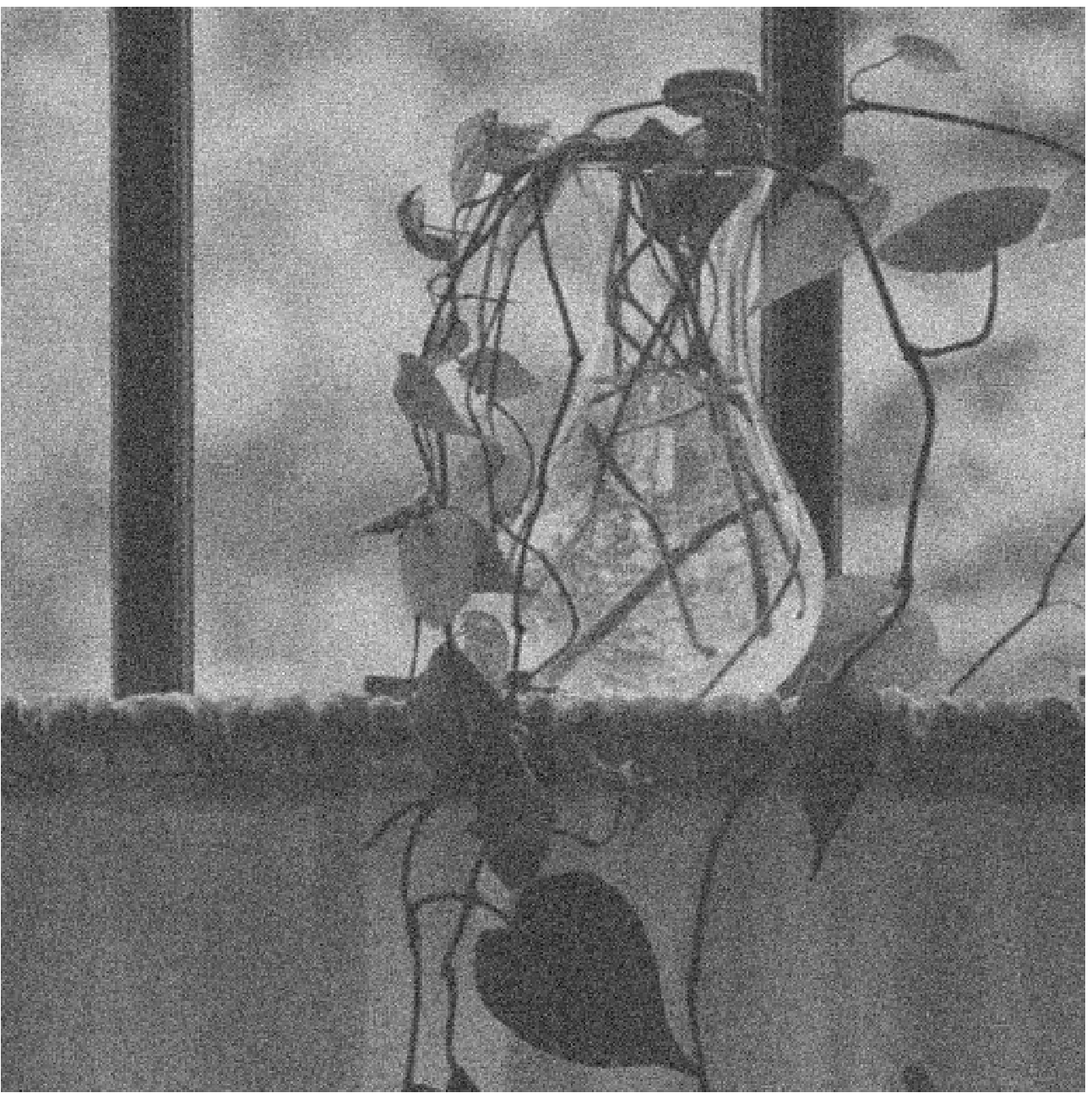}}
  \subfigure[OGSATV-ADM4]{
    \label{fig3:subfig:b}
    \includegraphics[width=0.23\textwidth]{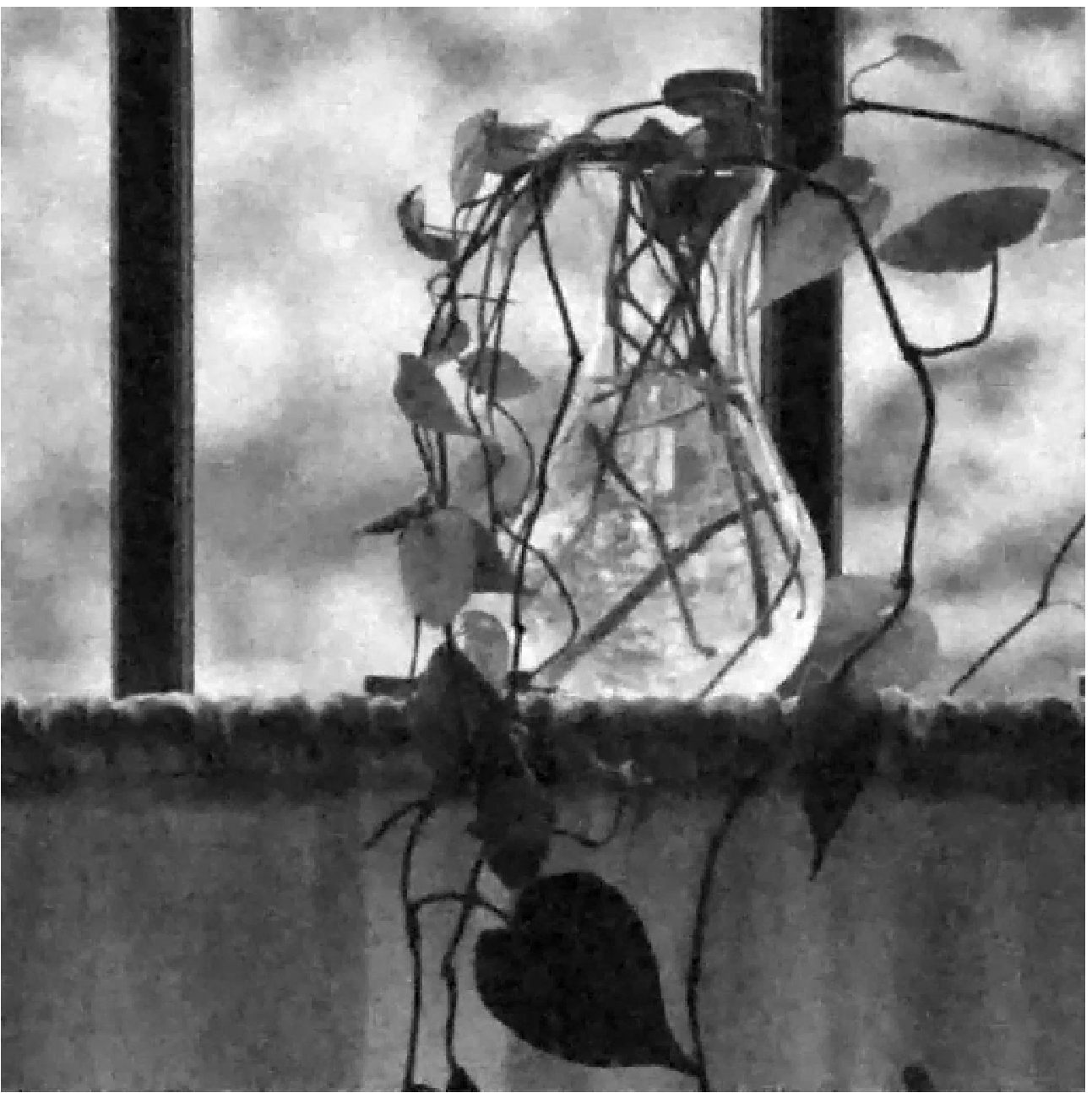}}
    \caption{Noisy ($\sigma$= 30) grayscale Vase image and the OGSATV-ADM4 estimate (PSNR 28.43 dB).}
\end{figure}
\ \\\\
$Example\ \textup{I}$: \textit{Image denoising}
\\

If the operator $H$ is an identity matrix, i.e., $H=I$, our goal is then to tackle the classic denoising problem. For this example, we compare our method with Chambolle's dual method \cite{Cham2004}, the split Bregman method \cite{GO2009} and LLT-ALM method \cite{WT2010}. Chambolle's dual method is a very famous nonlinear projection method proposed for image denoising. The split Bregman method proposed by Goldstein and Osher \cite{GO2009} for grayscale image denoising is well known for its high efficiency. The code was originally implemented in Matlab C/Mex Code by the authors, however, in the consideration of the fairness of comparisons, we re-implemented it in Matlab \verb"m-file" where Gauss-Seidel iteration was used to solve the its subproblem. In \cite{LLT2003}, Lysaker, Lundervold, and Tai proposed a representative high order TV models for overcoming staircase effect usually caused by TV models. This model is commonly called LLT model in the literature. From the perspective of implementing efficiency, Wu and Tai \cite{WT2010} adopted the augmented Lagrangian method (ALM) to solve this model. The implementation is also coded by us in Matlab.

\begin{figure}[htbp]
  \centering
  \subfigure[]{
    \label{fig4:subfig:a}
    \includegraphics[width=0.225\textwidth]{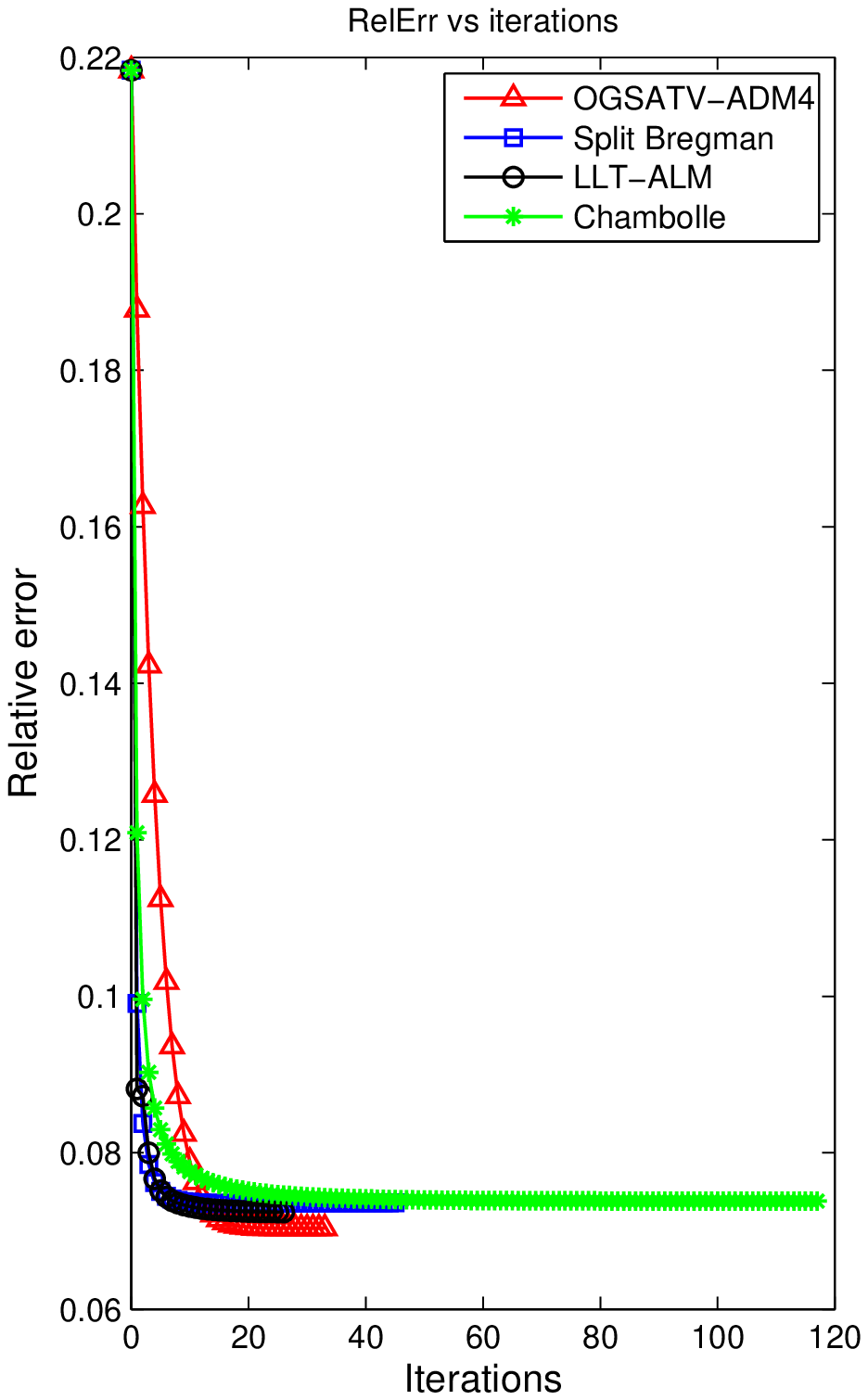}}
  \subfigure[]{
    \label{fig4:subfig:b}
    \includegraphics[width=0.22\textwidth]{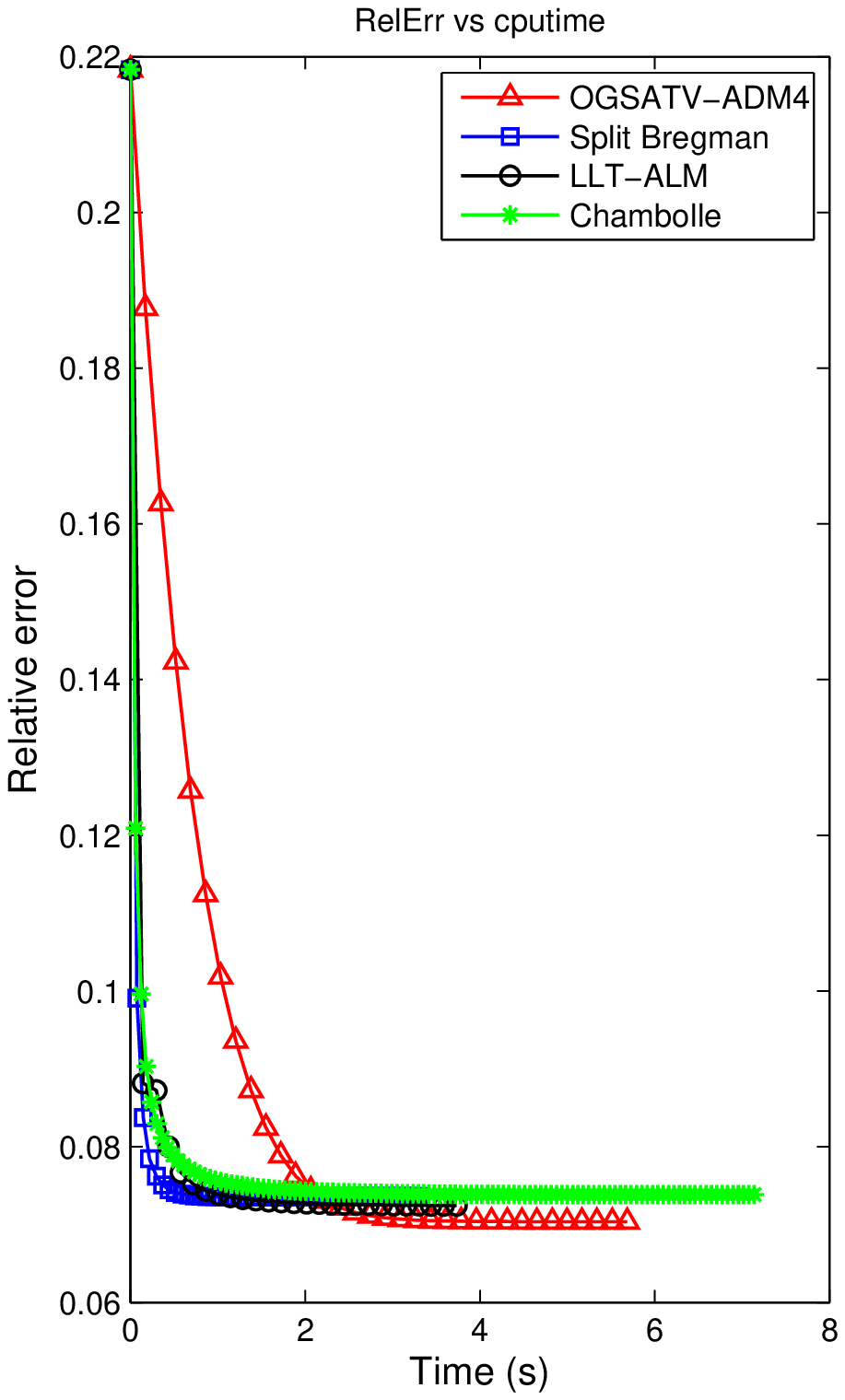}}
    \caption{ (a) Evolution of the RelErr along the iterations; (b) Evolution of the RelErr over elapsed time.}
\end{figure}
\begin{table*}[htbp]
\setlength{\abovecaptionskip}{0pt} \setlength{\belowcaptionskip}{12pt}
\centering
\caption{COMPARISON OF THE PERFORMANCE OF FOUR METHODS WITH DIFFERENT BLURRING KERNEL}
    \begin{tabular}{|r|r|c|c|c|c|}
    \hline
    \multicolumn{1}{|c|}{} &       & FastTV  &  CADMTVL2 & LLT-ALM & OGSATV-ADM4 \\
    \hline
    \multicolumn{1}{|c|}{Ker} & \multicolumn{1}{c|}{Image} & PSNR/RelErr/Time/Iter & PSNR/RelErr/Time/Iter & PSNR/RelErr/Time/Iter & PSNR/RelErr/Time/Iter \\
    \hline
    \multicolumn{1}{|c|}{\multirow{10}{*}{G}} & \multicolumn{1}{c|}{Barbara} & 27.48/0.079/4.70/88  & 27.48/0.079/0.30/19  & 28.15/0.073/1.85/61  & 28.28/0.072/0.70/13  \\
\cline{2-6}    \multicolumn{1}{|c|}{} & \multicolumn{1}{c|}{Boats} &  31.46/0.049/19.67/59   & 31.50/0.049/1.61/22  &  31.78/0.048/6.34/45 & 32.20/0.045/5.93/30   \\
\cline{2-6}    \multicolumn{1}{|c|}{} & \multicolumn{1}{c|}{Bridge} & 24.57/0.119/5.17/92  & 24.55/0.119/0.34/19 & 24.59/0.119/2.39/78 & 24.82/0.116/0.80/16  \\
\cline{2-6}    \multicolumn{1}{|c|}{} & \multicolumn{1}{c|}{C.man} & 28.60/0.071/4.12/63  & 28.61/0.071/0.50/23  & 28.12/0.075/1.77/53 & 28.82/0.069/0.97/20 \\
\cline{2-6}    \multicolumn{1}{|c|}{} & \multicolumn{1}{c|}{House} & 34.34/0.034/0.98/19 & 34.20/0.034/0.42/25 & 34.23/0.034/1.20/41  & 35.11/0.031/2.35/43 \\
\cline{2-6}    \multicolumn{1}{|c|}{} & \multicolumn{1}{c|}{Lena} & 30.22/0.070/2.93/53 & 30.24/0.070/0.39/25 & 30.19/0.071/1.58/41 & 30.63/0.067/1.89/30  \\
\cline{2-6}    \multicolumn{1}{|c|}{} & \multicolumn{1}{c|}{Peppers} & 31.68/0.050/2.29/44  &  31.46/0.051/0.37/24 & 31.36/0.051/2.02/65  & 32.14/0.047/1.57/30 \\
\cline{2-6}    \multicolumn{1}{|c|}{} & \multicolumn{1}{c|}{Butterfly} & 28.34/0.075/11.35/47 & 28.47/0.074/1.24/23 & 28.66/0.073/6.55/64 & 28.96/0.070/2.44/19 \\
\cline{2-6}    \multicolumn{1}{|c|}{} & \multicolumn{1}{c|}{W.station} & 30.61/0.060/23.35/81  & 30.62/0.060/1.70/23 & 30.84/0.059/7.88/58 & 31.36/0.056/6.18/30 \\
\cline{2-6}    \multicolumn{1}{|c|}{} & \multicolumn{1}{c|}{Jellyfish} & 35.90/0.031/15.85/49 & 36.15/0.030/1.95/27 &  37.25/0.026/3.47/26  & 37.59/0.025/6.36/38 \\
    \hline
    \hline
    \multicolumn{1}{|c|}{\multirow{10}{*}{A}} & \multicolumn{1}{c|}{Barbara} & 28.18/0.073/3.86/76 &  28.20/0.073/0.31/20  & 28.80/0.068/2.07/71 & 29.09/0.066/0.73/15 \\
\cline{2-6}    \multicolumn{1}{|r|}{} & \multicolumn{1}{c|}{Boats} & 31.73/0.048/15.24/46 & 31.79/0.048/1.65/23   & 32.00/0.046/7.13/50   &  32.35/0.045/4.44/23\\
\cline{2-6}    \multicolumn{1}{|r|}{} & \multicolumn{1}{c|}{Bridge} &  25.13/0.112/5.09/100  & 25.13/0.112/0.30/20  & 25.10/0.112/2.37/82 & 25.44/0.108/1.19/24  \\
\cline{2-6}    \multicolumn{1}{|r|}{} & \multicolumn{1}{c|}{C.man} &  29.32/0.065/1.08/22  & 29.41/0.064/0.37/24 & 28.59/0.071/1.51/54  & 29.45/0.064/1.03/26  \\
\cline{2-6}    \multicolumn{1}{|r|}{} & \multicolumn{1}{c|}{House} & 34.81/0.032/3.16/62 & 35.05/0.031/0.41/27  & 35.23/0.030/0.92/32 & 36.10/0.027/1.28/26 \\
\cline{2-6}    \multicolumn{1}{|r|}{} & \multicolumn{1}{c|}{Lena} & 30.54/0.068/2.56/51  &  30.55/0.068/0.35/23 & 30.35/0.069/1.42/49  & 30.77/0.066/0.72/15  \\
\cline{2-6}    \multicolumn{1}{|r|}{} & \multicolumn{1}{c|}{Peppers} & 32.13/0.047/1.95/39  & 32.16/0.047/0.38/25  & 31.84/0.049/1.56/54  & 32.52/0.045/1.14/23 \\
\cline{2-6}    \multicolumn{1}{|r|}{} & \multicolumn{1}{c|}{Butterfly} & 28.85/0.071/17.01/72 & 28.93/0.070/1.27/25 &  29.14/0.069/7.29/72  & 29.39/0.067/4.39/31 \\
\cline{2-6}    \multicolumn{1}{|r|}{} & \multicolumn{1}{c|}{W.station} & 31.31/0.056/14.63/57  & 31.23/0.056/1.51/23  & 31.40/0.055/9.14/72 & 32.19/0.050/7.01/36  \\
\cline{2-6}          & \multicolumn{1}{c|}{Jellyfish} & 35.61/0.032/19.35/64  & 35.84/0.031/1.98/29 & 36.60/0.029/3.19/25  & 36.68/0.028/6.46/38 \\
    \hline
    \end{tabular}%
  \label{tab:addlabel}%
\end{table*}%
We make use of 9 test images for comparisons. We simulate the noisy images with two different noise levels. All of these images were corrupted by the zero-mean additive Gaussian noise with the standard deviation $\delta = 30$ and $\delta = 15$, respectively.

In Fig. 3, we show a noisy ($\delta=30$) Vase image and the corresponding denoised version using proposed OGSATV-ADM4. Visually, we see that OGSATV-ADM4 works very well for image denoising. The evolutions of RelErrs vs iterations and CPU time using four different methods are plotted in Fig. 4. From this figure, we observe that Split Bregman method converges extremely fast and consumes the least CPU time, and Chambolle's method is the slowest one no matter in terms of iterations or consuming time. However, our method OGSATV-ADM4 reach the lowest RelErr value at the convergence point in reasonable time.

Moreover, the denoising comparison among four different methods is further illustrated in Fig. 5, where we show fragments of two true images, noisy images ($\delta = 30$) and the corresponding denoised ones. As can be seen from the Fig. 5, denoised images obtained by using TV methods (the split Bregman method and Chambolle's dual method) have apparent staircase effect (such as the parts pointed by the left below arrow and upper right arrow of Boats image, and the nose and the lower jaw of Man image), while the LLT-ALM method and the OGSATV-ADM4 overcome this drawback to a great extent. However, there also exists shortcoming caused by LLT-ALM, i.e., some parts of the restored images are overly smoothed. The parts pointed by the upper left arrow and lower right arrow of Boats image, and the eyelid and lips (the parts pointed by left two arrows) of Man image show this effect. Note that our method can avoid this drawback effectively.
\begin{figure*}[!ht]
  \centering
  \subfigure[True image]{
    \label{fig5:subfig:a}
    \includegraphics[width=1.0in]{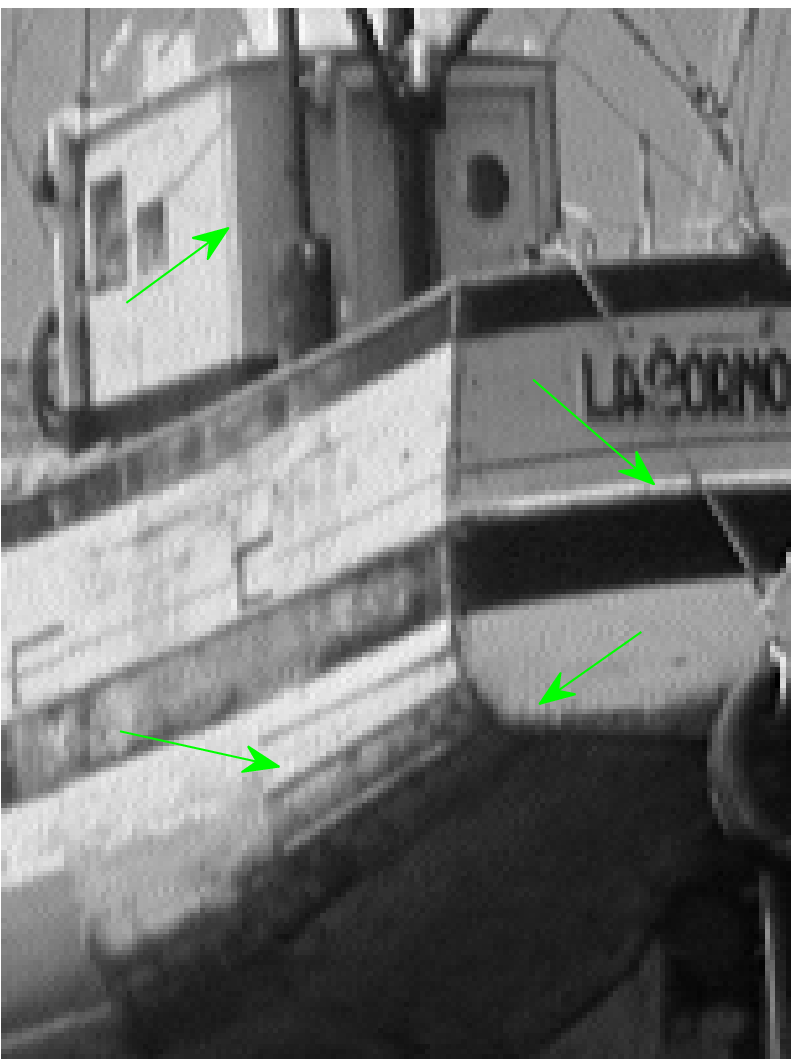}}
  \subfigure[Noisy image]{
    \label{fig5:subfig:b}
    \includegraphics[width=1.0in]{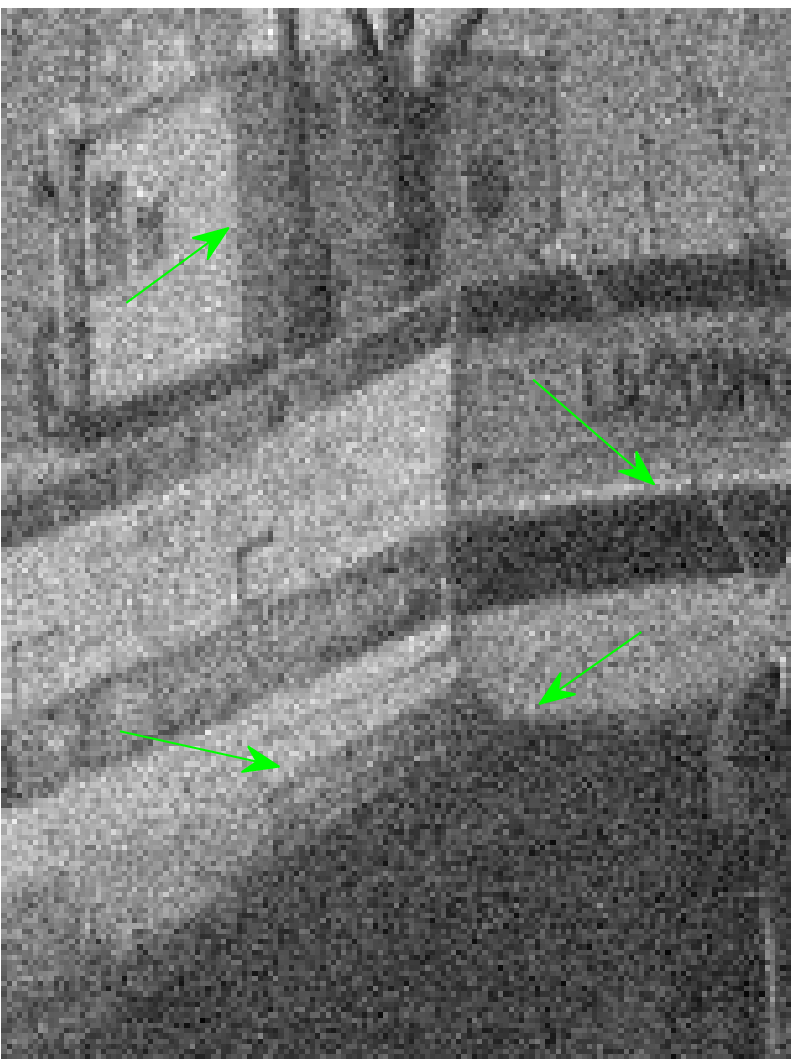}}
  \subfigure[Split Bregman]{
    \label{fig5:subfig:c}
    \includegraphics[width=1.0in]{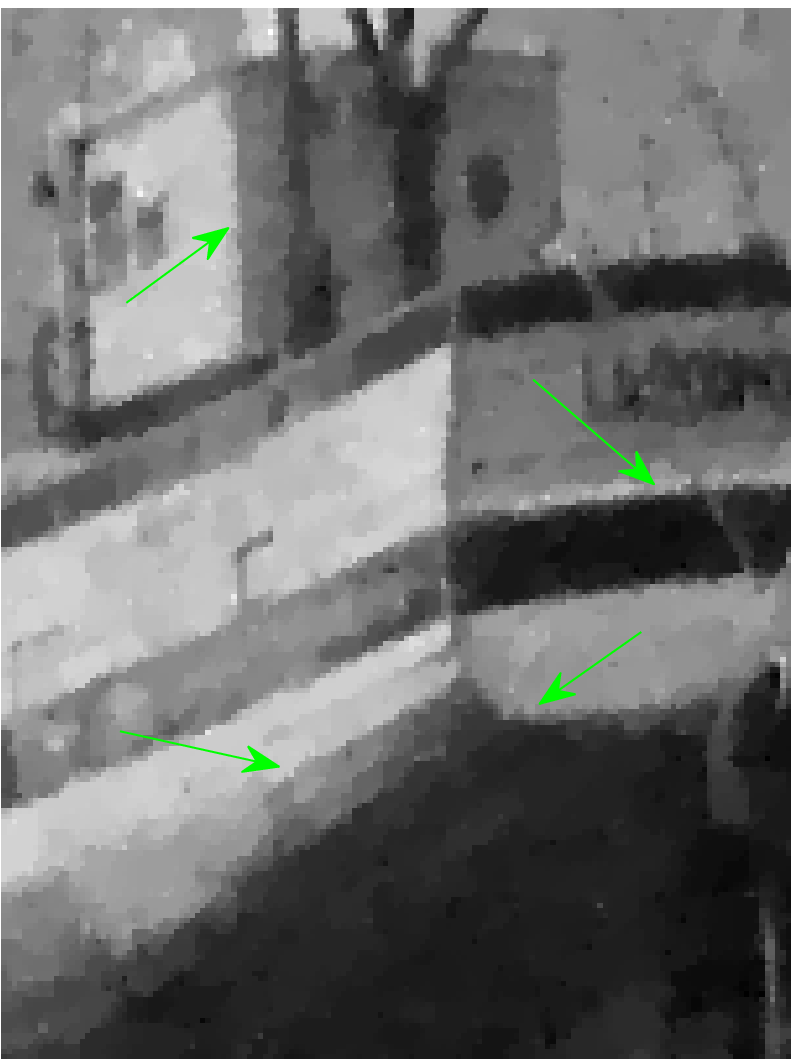}}
    \subfigure[Chambolle]{
    \label{fig5:subfig:d}
    \includegraphics[width=1.0in]{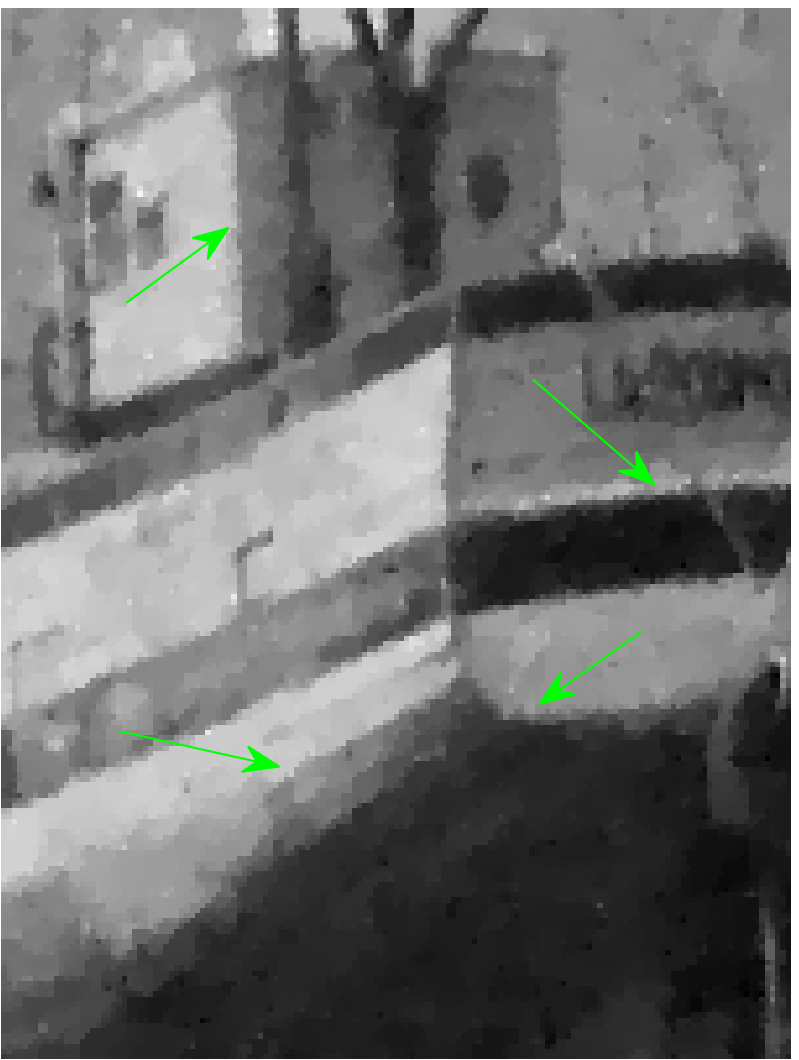}}
  \subfigure[LLT-ALM]{
    \label{fig5:subfig:e}
    \includegraphics[width=1.0in]{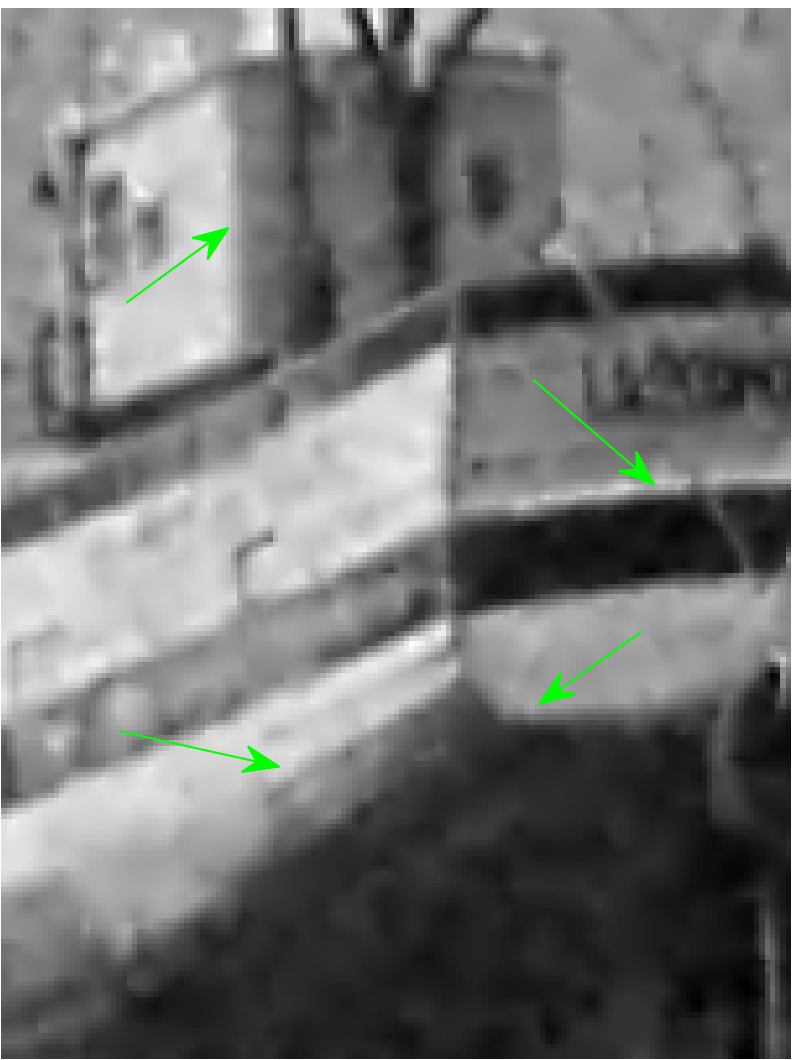}}
  \subfigure[OGSATV-ADM4]{
    \label{fig5:subfig:f}
    \includegraphics[width=1.0in]{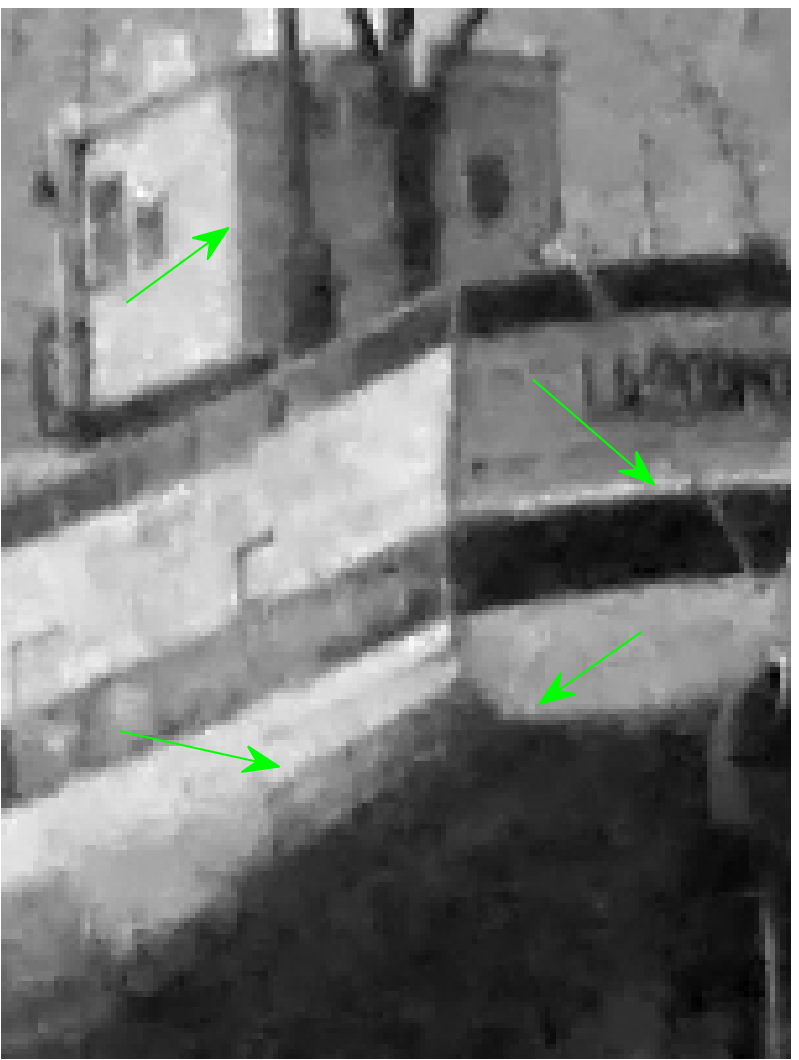}}
\\
  \subfigure[True image]{
    \label{fig5:subfig:g}
    \includegraphics[width=1.0in]{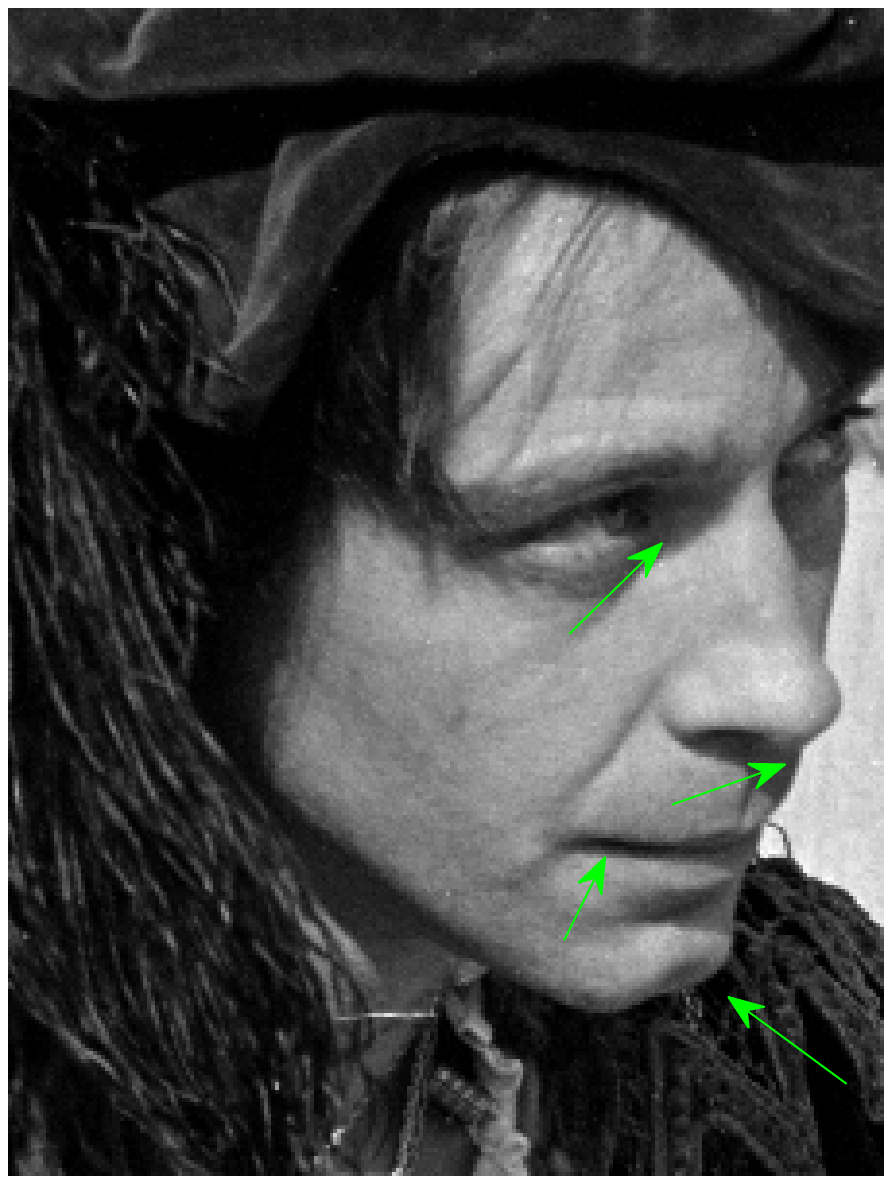}}
  \subfigure[Noisy image]{
    \label{fig5:subfig:h}
    \includegraphics[width=1.0in]{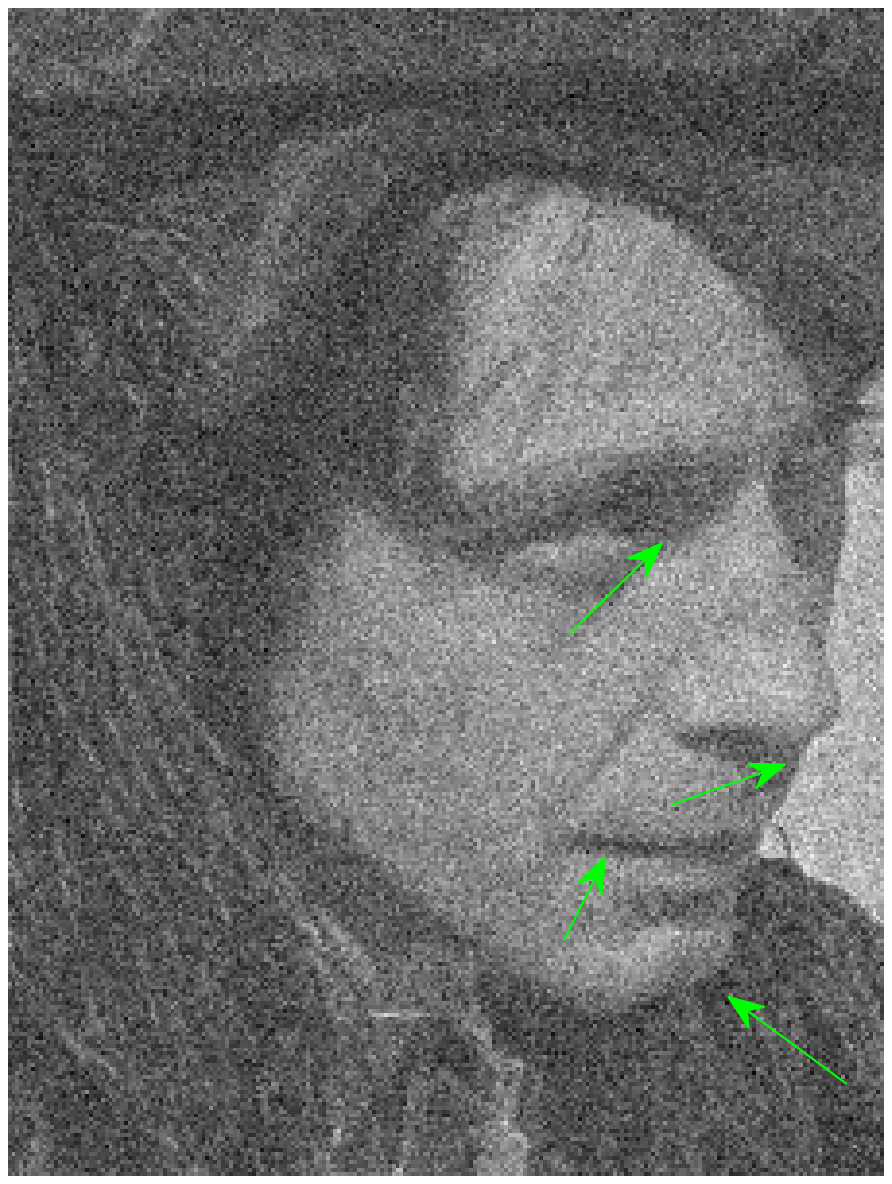}}
  \subfigure[Split Bregman]{
    \label{fig5:subfig:i}
    \includegraphics[width=1.0in]{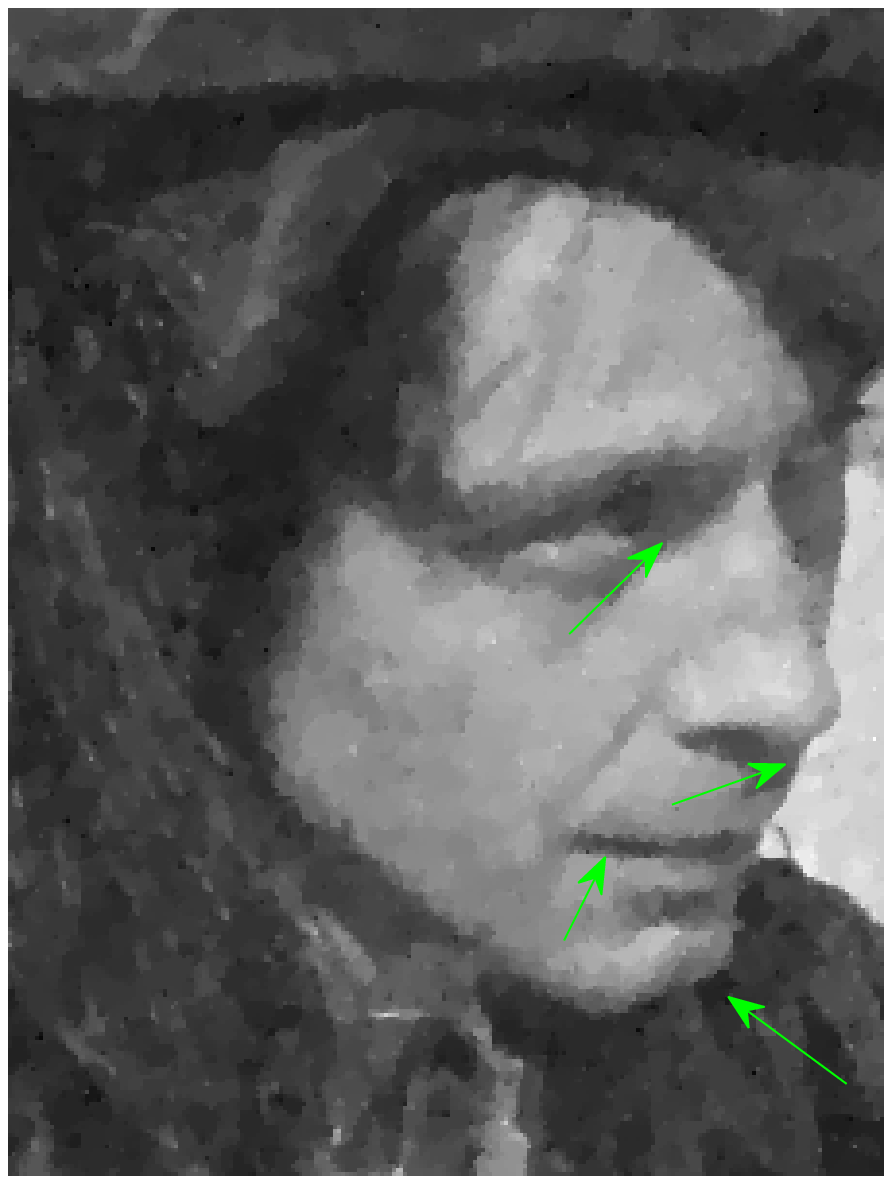}}
    \subfigure[Chambolle]{
    \label{fig5:subfig:j}
    \includegraphics[width=1.0in]{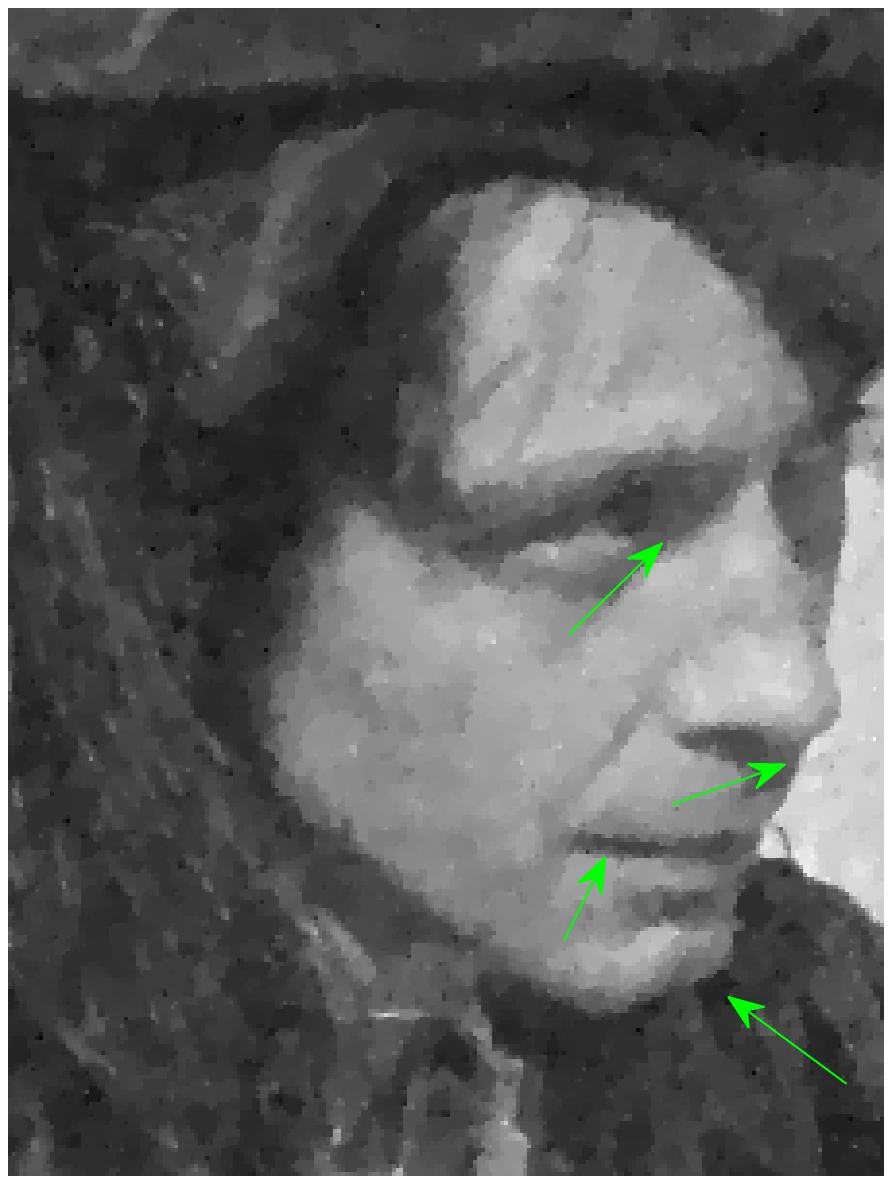}}
  \subfigure[LLT-ALM]{
    \label{fig5:subfig:k}
    \includegraphics[width=1.0in]{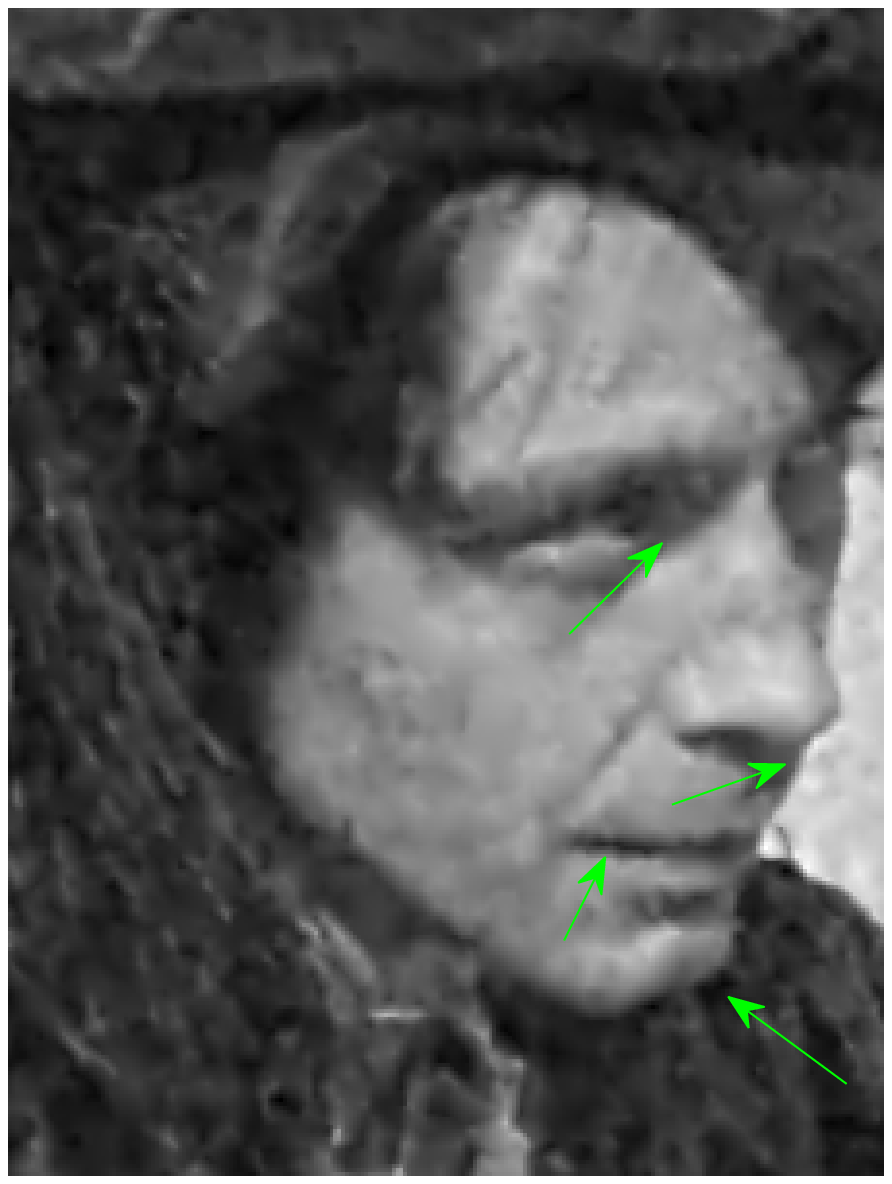}}
  \subfigure[OGSATV-ADM4]{
    \label{fig5:subfig:l}
    \includegraphics[width=1.0in]{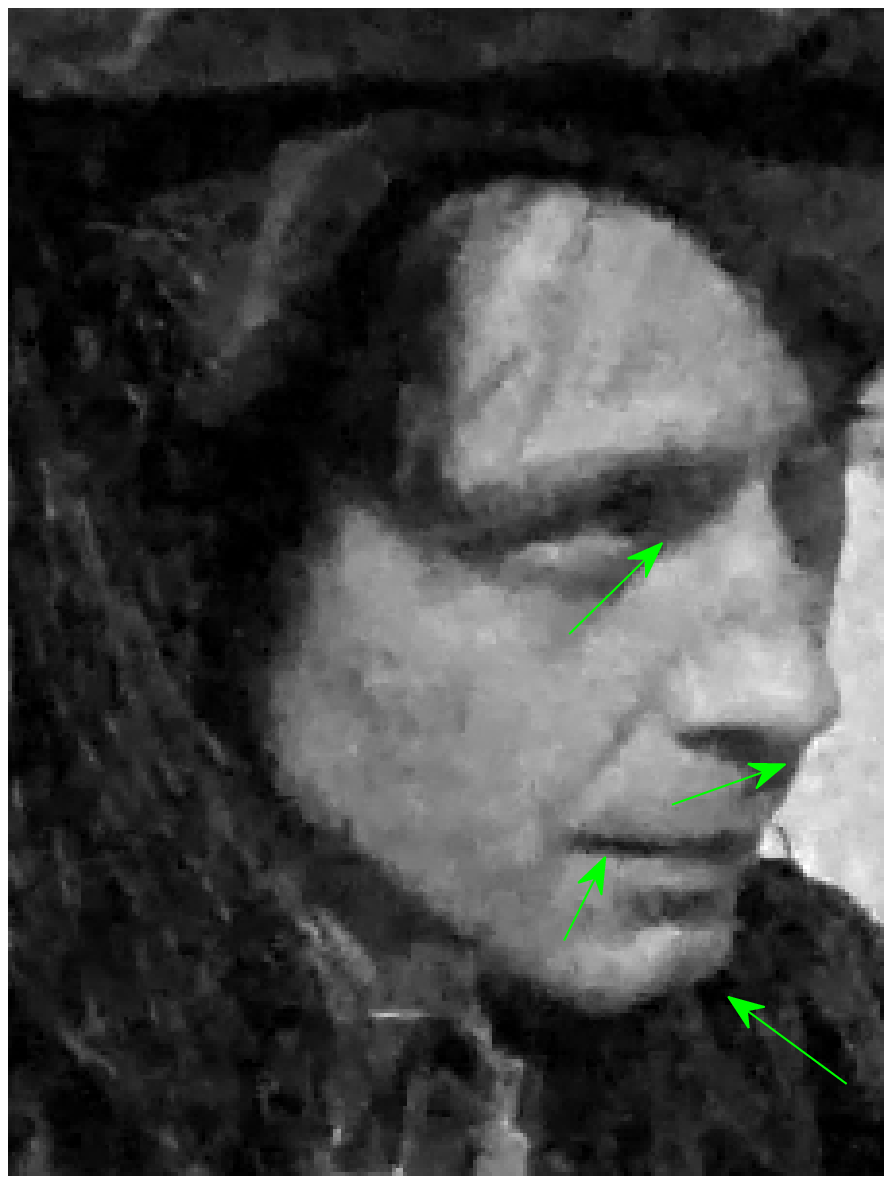}}
\caption{Fragments of the grayscale Boats (top row) and Man (bottom row) denoised by (from left to right): Split Bregman method \cite{GO2009}, Chambolle's method \cite{Cham2004}, LLT-ALM \cite{WT2010} and the proposed OGSATV-ADM4 for denoising with $\delta$= 30.}
\end{figure*}
The output results in terms of PSNR, RelErr, CPU time, and iterations of four methods are given in Table II. From the table, we observe that the split Bregman method and Chambolle's dual method achieve similar PSNR results, while the LLT-ALM method can sometimes performs better than both of them in terms of PSNR. Overall, our method OGSATV-ADM4 can reach the highest PSNR results among the four methods. We should also note that the split Bregman usually costs least CPU time.

\ \\\\
$Example\ \textup{II}$: \textit{Image deblurring}
\\

In case $H$ is a blurring matrix, then the problem we aim to solve is deblurring. For this example, we compare our method with the method FastTV  proposed by Huang, Ng, and Wen \cite{HNW2008}, the ADMM method for solving constrained TV-L2 model (CADMTVL2) by Chan, Tao, and Yuan \cite{CTY2013} and the method LLT-ALM \cite{WT2010}. Note that the test images used in CADMTVL2 \cite{CTY2013} are all scaled to the interval $[0, 1]$, so the box constraint in their constrained models is simply $[0, 1]$.

\begin{figure}[!ht]
  \centering
  \subfigure[Blurred image]{
    \label{fig6:subfig:a}
    \includegraphics[width=0.225\textwidth]{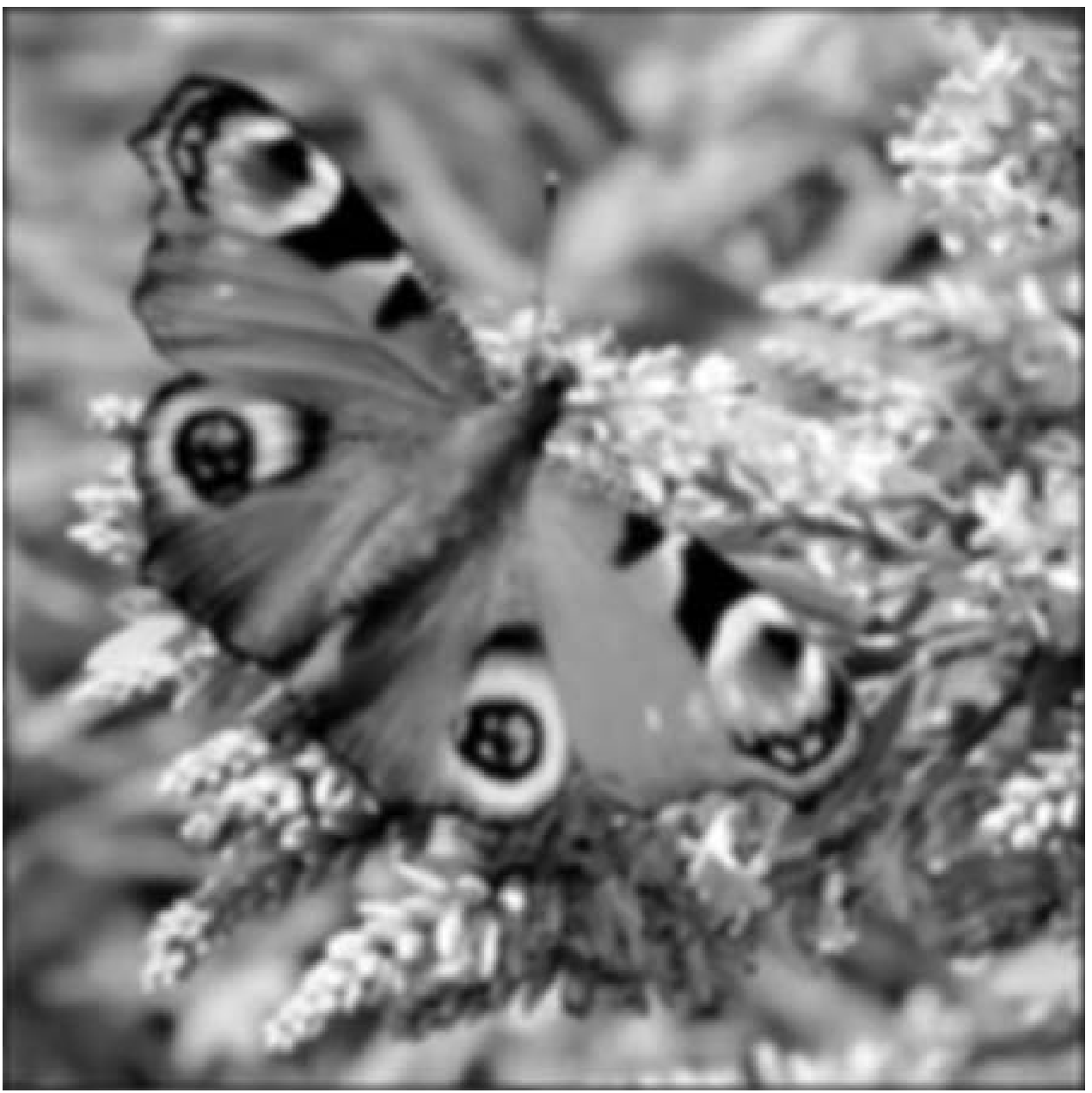}}
  \subfigure[OGSATV-ADM4]{
    \label{fig6:subfig:b}
    \includegraphics[width=0.225\textwidth]{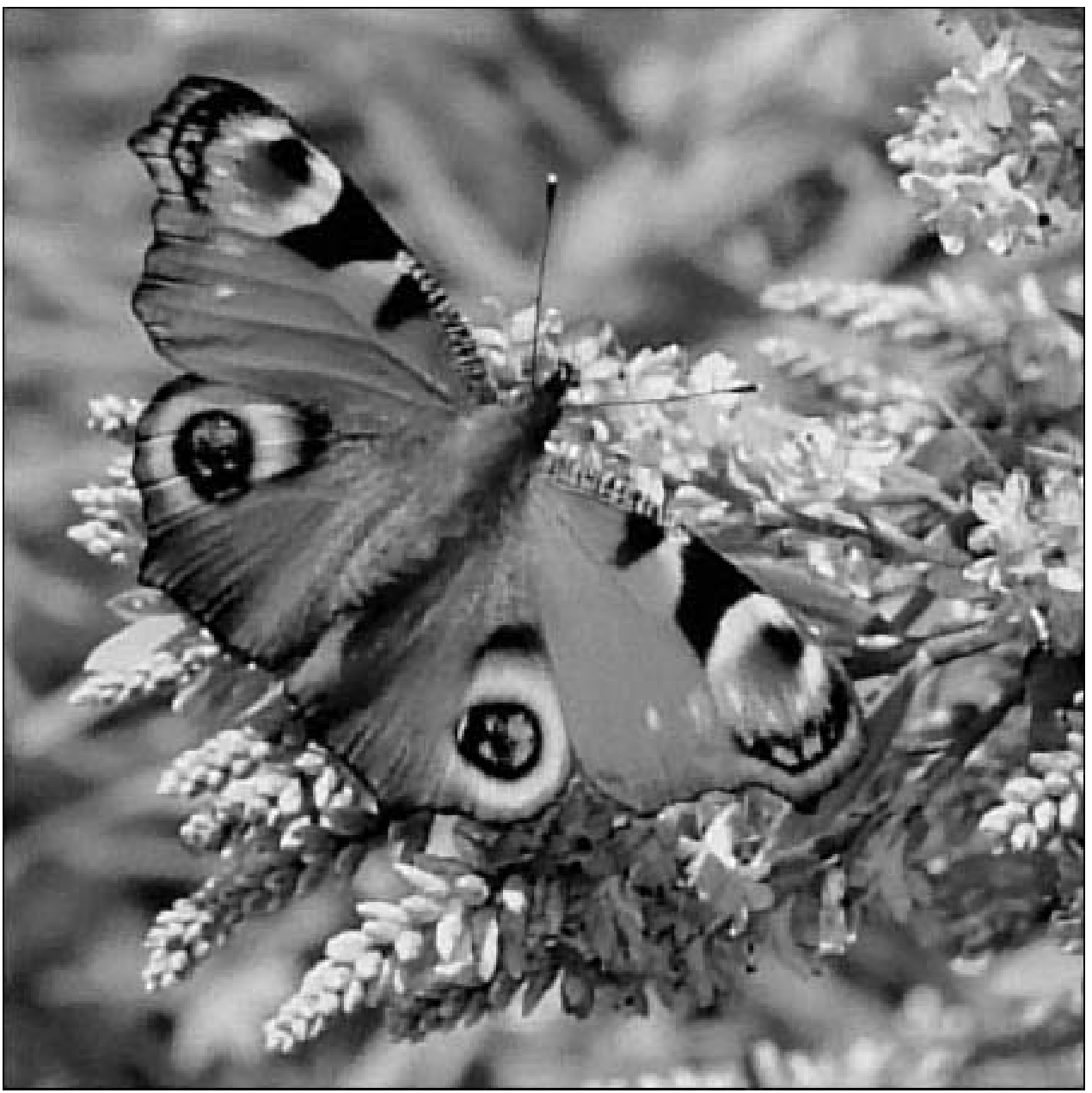}}\\
  \subfigure[Blurred image]{
    \label{fig6:subfig:c}
    \includegraphics[width=0.225\textwidth]{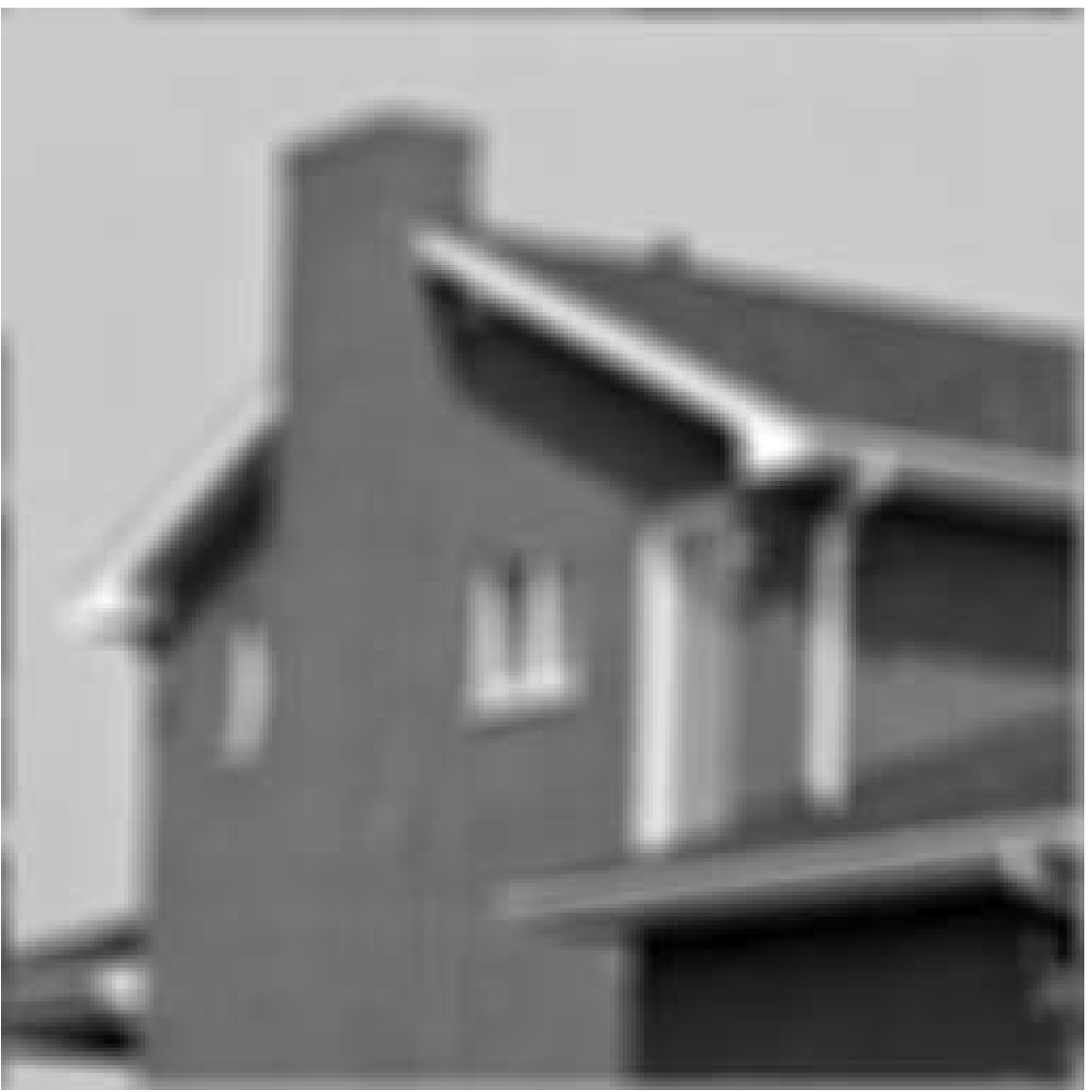}}
  \subfigure[OGSATV-ADM4]{
    \label{fig6:subfig:d}
    \includegraphics[width=0.225\textwidth]{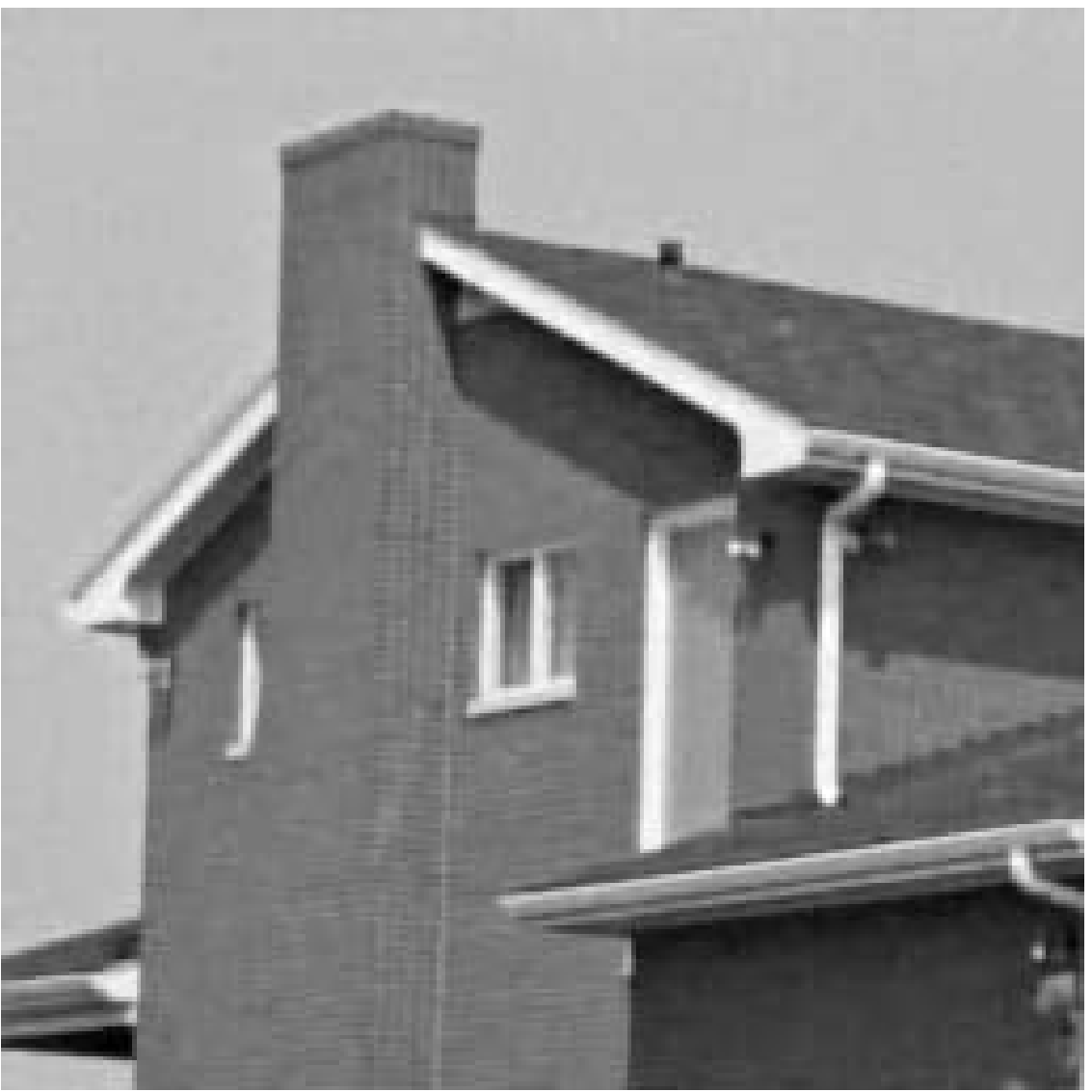}}
\caption{Restorations of the ``Butterfly" image with Gaussian blur and the ``House" image with average blur.}
\end{figure}

\begin{figure}[!ht]
  \centering
  \subfigure[]{
    \label{fig7:subfig:a}
    \includegraphics[width=0.22\textwidth]{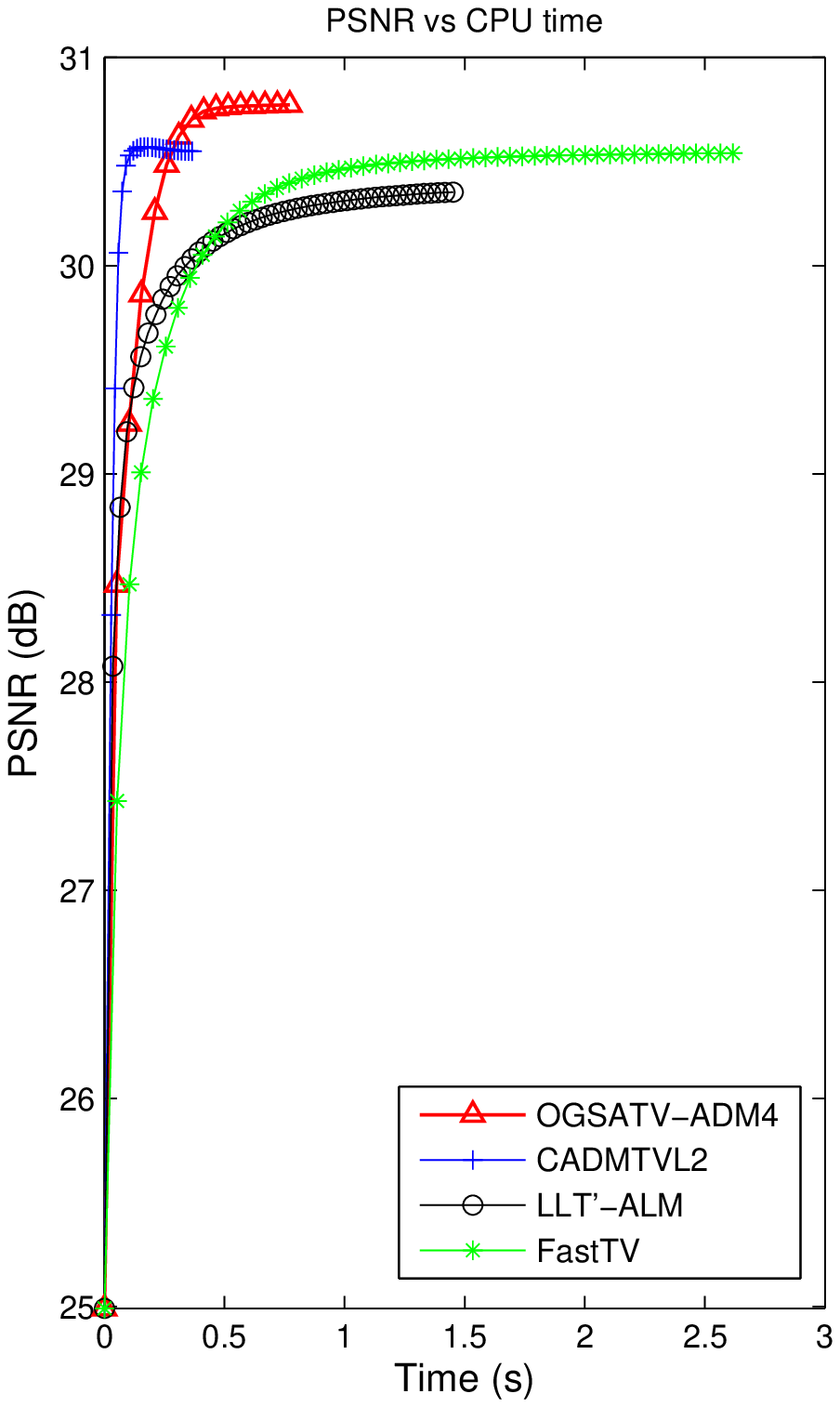}}
  \subfigure[]{
    \label{fig7:subfig:b}
    \includegraphics[width=0.22\textwidth]{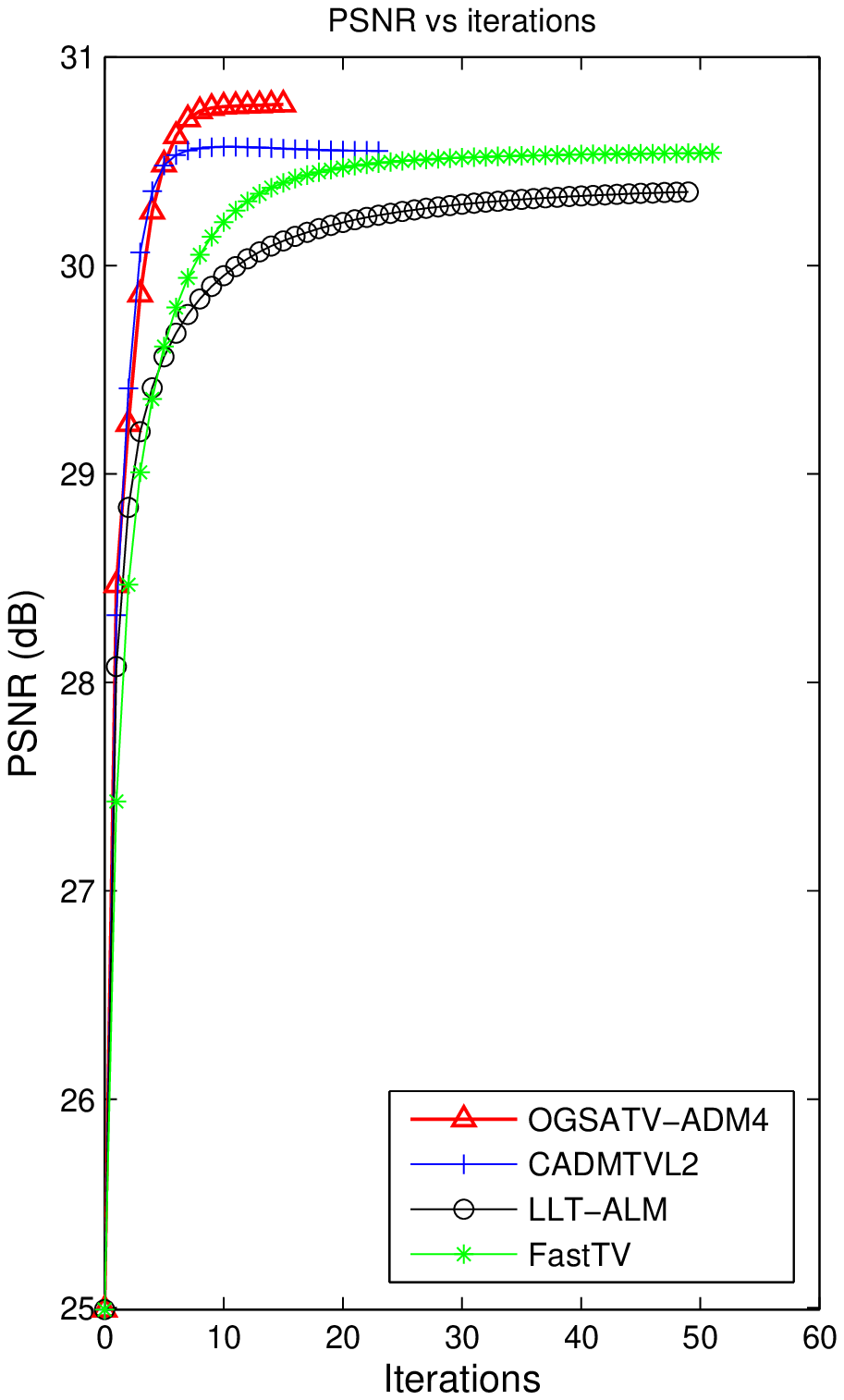}}
\caption{Restoration of the ``Lena" image with average blur: evolution of the PSNR over time and iterations.}
\end{figure}

In this example, we test two different types of blurring kernels: Gaussian blur (G)  and average blur (A), which can be generated by the Matlab built-in function \verb"fspecial", more specifically, \verb"fspecial('Gaussian',[7 7],2)" and \verb"fspecial('average',9)" . For each blurring case, the blurred images are further corrupted by zero mean Gaussian noise with BSNR = 40. Two image estimates obtained by OGSATV-ADM4 are shown in Fig. 6, with the blurred images also shown for illustration. It is clear from Fig. 6 that the proposed method can restore blurred images effectively and in high quality. Fig. 7 shows the evolution of the PSNR over computing time and iterations for four different methods with respect to restoration of the ``Lena" image with average blur. It is obvious that our method reaches the highest PSNR with least iterations. It is obvious that CADMTVL2 needs fewer computing time to achieve the convergency point. We also observe that the penalty method FastTV needs the maximum computing time and iterations comparing with other three methods.

Table III shows the output results in terms of PSNR, RelErr, CPU time and iterations of four methods. From the table, we see that the quality in terms of PSNR of the restored images by FastTV and CADMTVL2 is almost the same. However, CADMTVL2 consumes much less time and needs much fewer iterations than FastTV. Overall, the proposed method reaches the highest PSNR compared with other three state-of-the-art methods, and needs less computing time and iterations than FastTV and LLT-ALM (except in the case of the ``Jellyfish" image, the computing time and iterations by using LLT-ALM are less than our method). Quantitatively, however, our method can obtain $0.3\sim1$ dB improvement in PSNR on average.
\begin{figure*}[!ht]
  \centering
  \subfigure[True image]{
    \label{fig8:subfig:a}
    \includegraphics[width=1.0in]{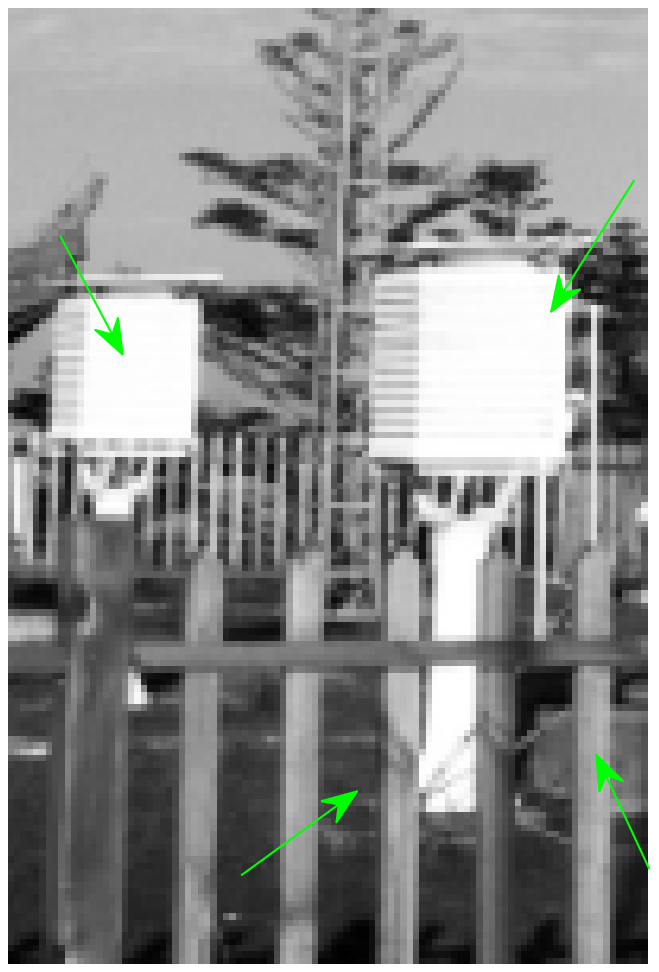}}
  \subfigure[Noisy image]{
    \label{fig8:subfig:b}
    \includegraphics[width=1.0in]{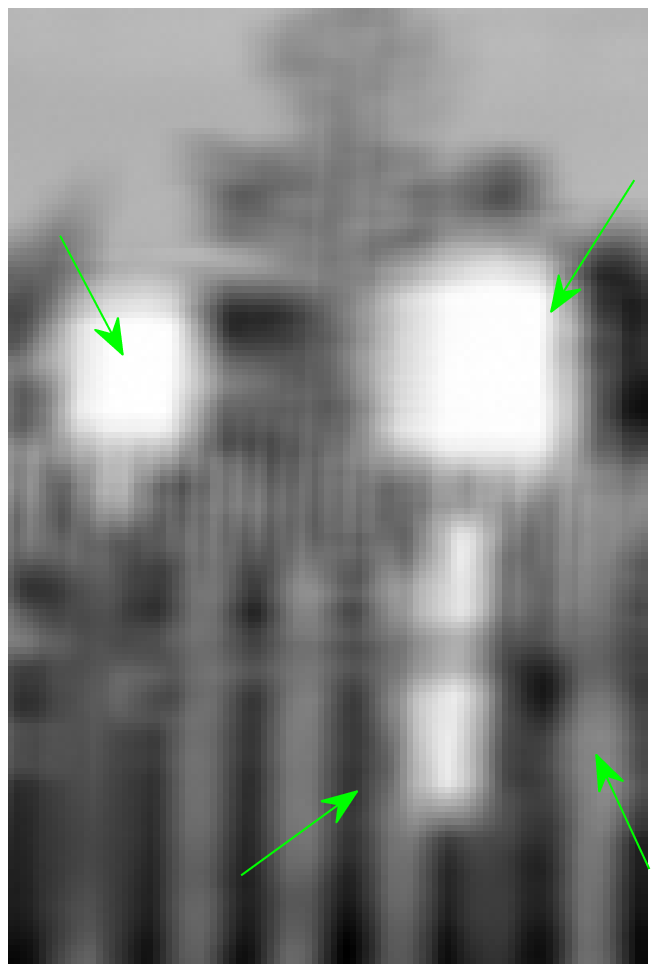}}
  \subfigure[FastTV]{
    \label{fig8:subfig:c}
    \includegraphics[width=1.0in]{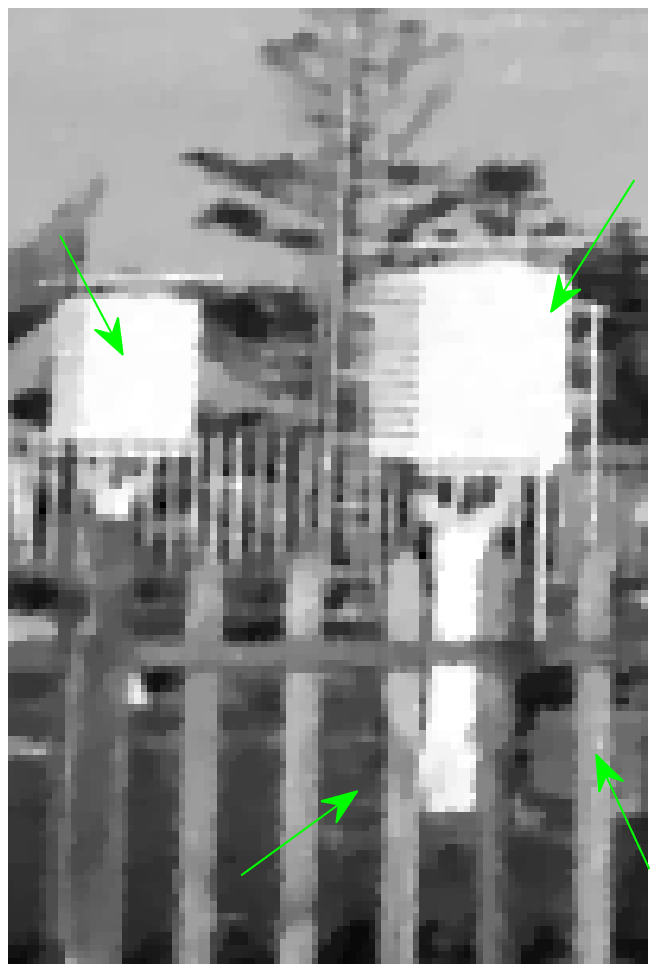}}
    \subfigure[CADMTVL2]{
    \label{fig8:subfig:d}
    \includegraphics[width=1.0in]{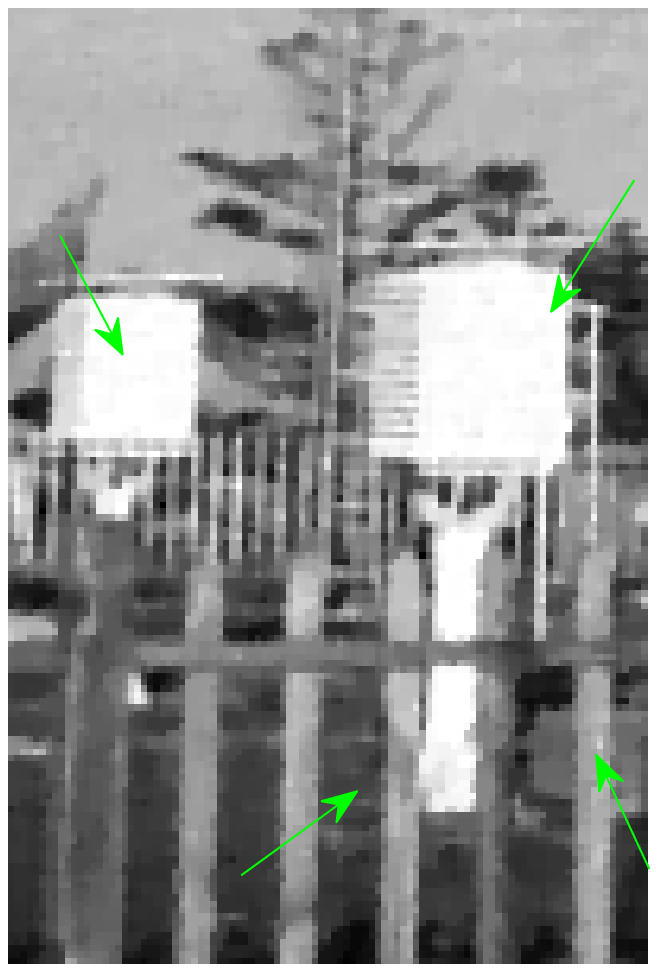}}
  \subfigure[LLT-ALM]{
    \label{fig:subfig:e}
    \includegraphics[width=1.0in]{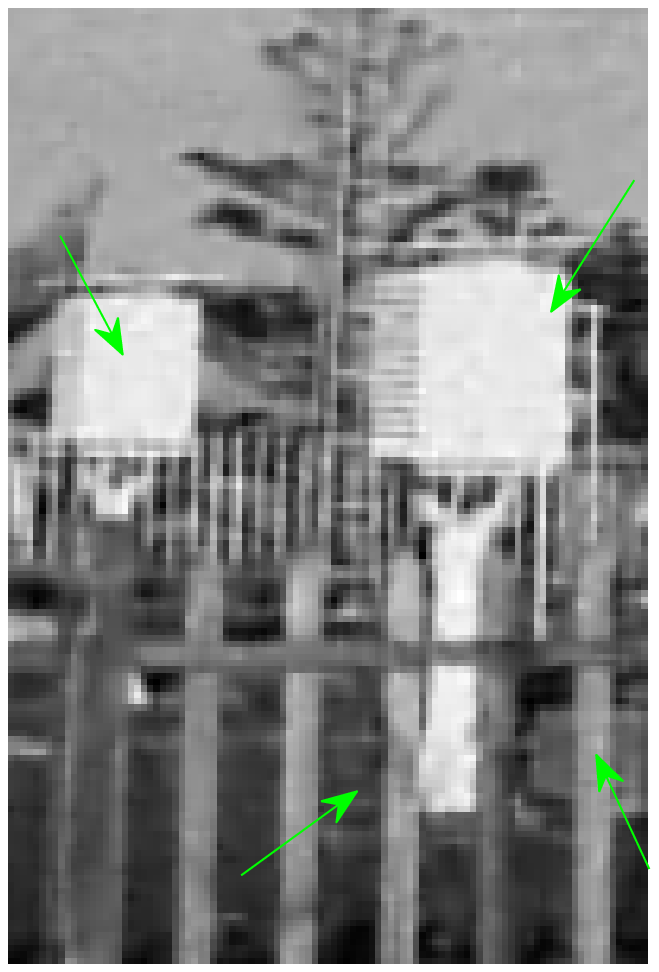}}
  \subfigure[OGSATV-ADM4]{
    \label{fig:subfig:f}
    \includegraphics[width=1.0in]{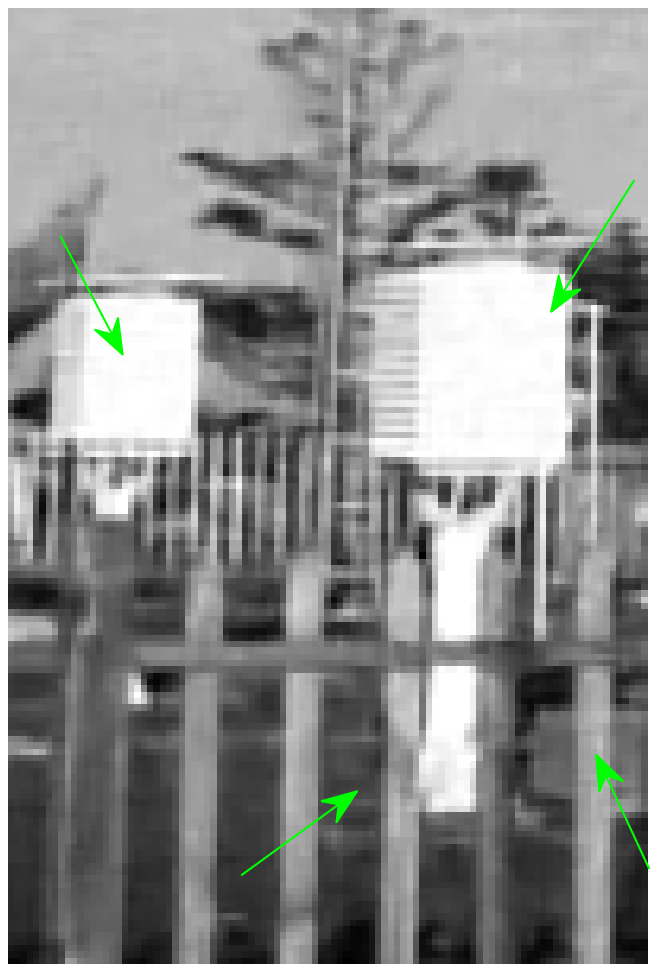}}
\\
  \subfigure[True image]{
    \label{fig8:subfig:g}
    \includegraphics[width=1.0in]{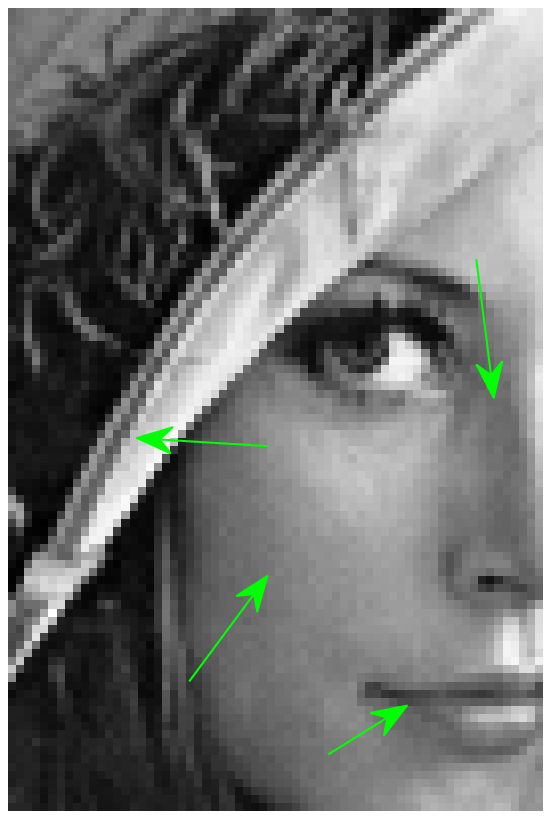}}
  \subfigure[Noisy image]{
    \label{fig8:subfig:h}
    \includegraphics[width=1.0in]{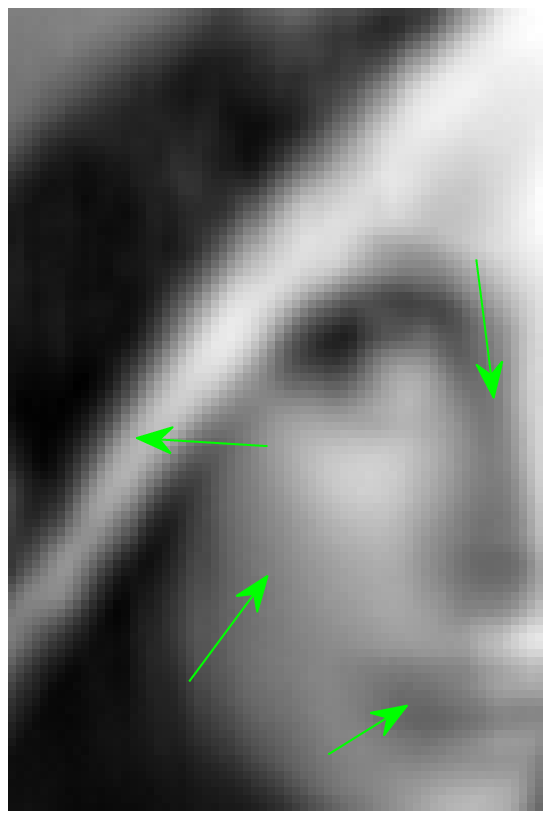}}
  \subfigure[FastTV]{
    \label{fig8:subfig:i}
    \includegraphics[width=1.0in]{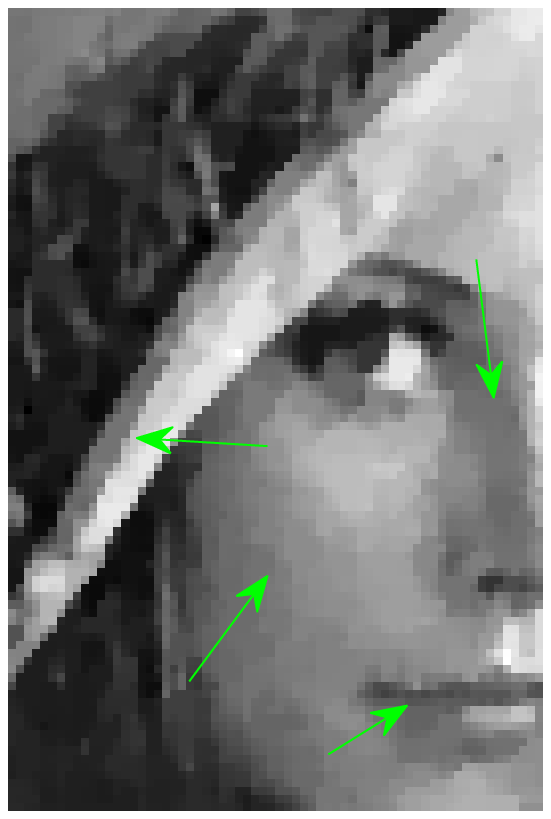}}
    \subfigure[CADMTVL2]{
    \label{fig8:subfig:j}
    \includegraphics[width=1.0in]{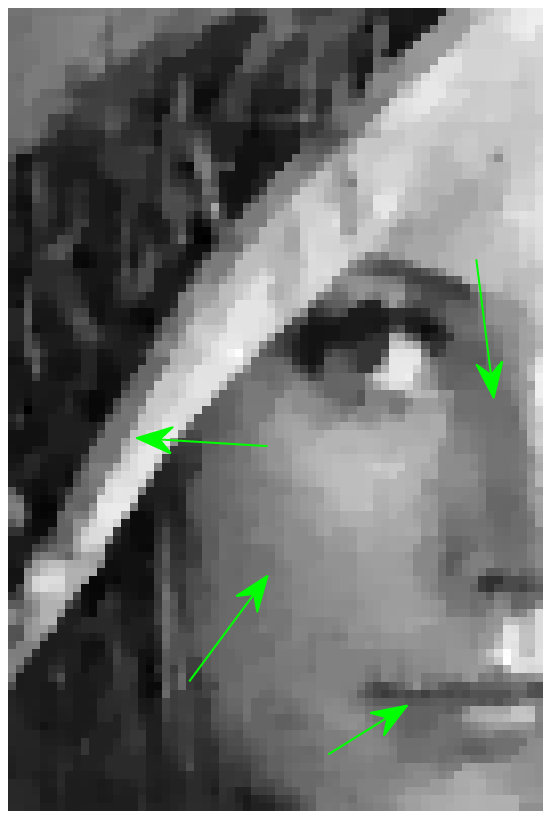}}
  \subfigure[LLT-ALM]{
    \label{fig8:subfig:k}
    \includegraphics[width=1.0in]{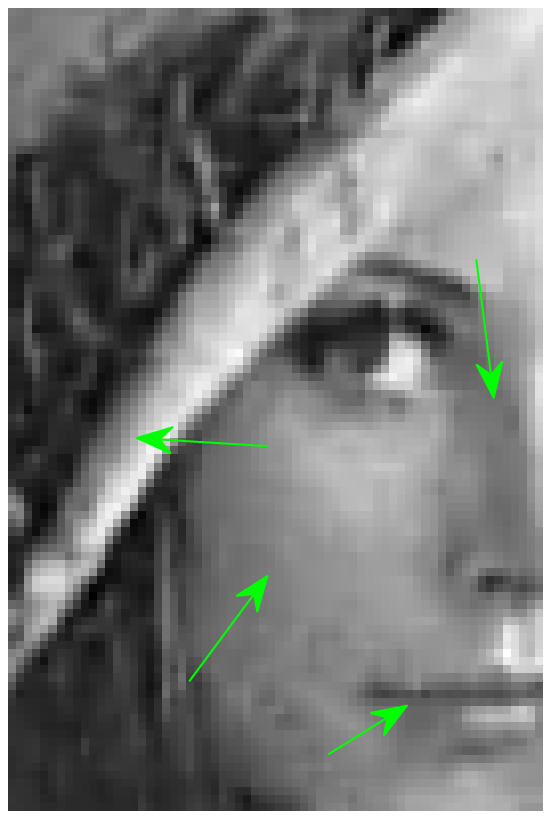}}
  \subfigure[OGSATV-ADM4]{
    \label{fig8:subfig:l}
    \includegraphics[width=1.0in]{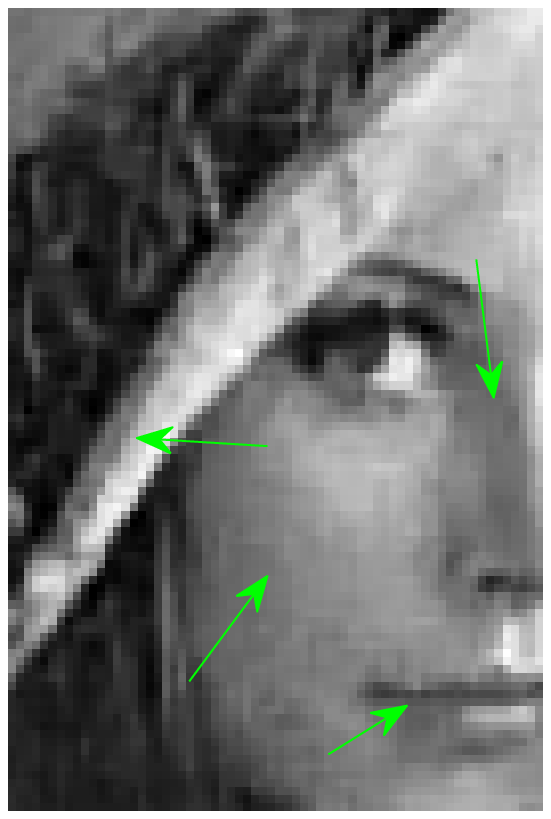}}

\caption{Fragments of the grayscale W.sation (top row) and Lena (bottom row) restored by (from left to right): FastTV \cite{HNW2008}, CADMTVL2 \cite{CTY2013}, LLT-ALM \cite{WT2010} and the proposed OGSATV-ADM4 (with average blur, BSNR= 40).}
\end{figure*}

Moreover, in order to illustrate the superior capability of our method for image deblurring. We show the fragments of restored images ``W.station" and ``Lena" in Fig. 8. In the top row of Fig. 8, we observe that the fences of the W.station image estimates obtained by FastTV and CADMTVL2 are very blocky (staircase effect), however, they are restored very well by both LLT-ALM and the proposed method. On the other hand, LLT-ALM makes the white boxes locally over-smoothed while our method, together with FastTV and CADMTVL2, can restore them almost the same as the true image. Similar phenomena can also be seen from the bottom row of Fig. 8, LLT-ALM and the proposed method can avoid staircase effect effectively, such as the lips and cheek. However, we notice that LLT-ALM fails to recover the brim of the hat (edges) correctly since it makes the brim over-smoothed. In contrast, our method can not only recover the edges very well, but avoids staircase effect as well.

\section{Conclusion}
In this paper, we study the image restoration problem based on the overlapping group sparsity total variation regularizer. To solve the corresponding minimization problem, we proposed a very efficient algorithm OGSATV-ADM4 under the framework of the classic ADMM and using MM method to tackle the associated subproblem. The numerical comparisons with many state-of-the-art methods show that our method is very effective and efficient. The results verify that the proposed method avoids staircase effect and yet preserves edges.

We are currently working on extending our method to real applications involving  compressed sensing, blind deconvolution, image enhancement and so on.
\section*{Acknowledgment}
The authors would like to thank Prof. M. Tao for providing us the code (CADMTVL2) in \cite{CTY2013} and Prof. M. Ng for making their code (FastTV) in \cite{HNW2008} available online.

\end{document}